\documentclass[twocolumn]{svjour3}          % twocolumn
\smartqed  % flush right qed marks, e.g. at end of proof
\usepackage{graphicx}
\usepackage{amsfonts}
\usepackage{amsmath}
\usepackage{subcaption}

\usepackage{todonotes}

\usepackage{color}
\usepackage{amssymb}
\usepackage[showonlyrefs]{mathtools}

\usepackage{hyperref}

\DeclareMathOperator{\supp}{supp}

\newcommand{\cE}{\mathcal{E}}
\newcommand{\cI}{\mathcal{I}}

\newcommand{\bP}{\mathbb{P}}

\newcommand{\cU}{\mathcal{U}}

\newcommand{\R}{\mathbb{R}}

\newcommand{\be}{\begin{equation}}
\newcommand{\ee}{\end{equation}}

\newcommand{\bS}{\mathbb{S}}

\DeclareMathOperator{\bR}{\mathbb{R}}

\usepackage[normalem]{ulem}

 \journalname{Journal of Mathematical Imaging and Vision}
\begin{document}

\title{Cortical-inspired Wilson-Cowan-type equations for orientation-dependent contrast perception modelling}
%\thanks{LC, VF and DP acknowledge the support of a public grant overseen by the French National Research Agency (ANR) as part of the \emph{Investissement d'avenir program}, through the iCODE project funded by the IDEX Paris-Saclay, ANR-11-IDEX-0003-02 and of the research project \emph{LiftME} funded by INS2I, CNRS.  VF and DP also acknowledge the support of ANR-15-CE40-0018 project \textit{SRGI - Sub-Riemannian Geometry and Interactions}. VF acknowledges the support received from the European Union's Horizon 2020 research and innovation programme under the Marie Sklodowska-Curie grant agreement No 794592. BF acknowledges the support of the Fondation Asile des Aveugles. }
\titlerunning{Cortical-inspired Wilson-Cowan modelling for orientation-dependent contrast perception}

% \subtitle{Do you have a subtitle?\\ If so, write it here}

\author{Marcelo Bertalm\'io
\and
Luca Calatroni
\and
Valentina Franceschi
\and
Benedetta Franceschiello
\and
 Dario Prandi 
 }

%\authorrunning{Short form of author list} % if too long for running head

\institute{Marcelo Bertalm\'io\at
DTIC, Universitat Pompeu Fabra, Barcelona, Spain\\
\href{mailto:marcelo.bertalmio@upf.edu}{marcelo.bertalmio@upf.edu}
\and
Luca Calatroni\at
{Universit\'e C\^{o}te d’Azur, CNRS, Inria, I3S, France \\
\href{mailto:calatroni@i3s.unice.fr}{calatroni@i3s.unice.fr}}
\and
Valentina Franceschi\at
LJLL, Sorbonne Universit\'e, Paris, France\\
\href{mailto:franceschiv@ljll.math.upmc.fr}{franceschiv@ljll.math.upmc.fr}
\and
Benedetta Franceschiello\at
 Fondation Asile des Aveugles and LINE, Dep. of Radiology, CHUV, Lausanne, Switzerland\\
\href{mailto:benedetta.franceschiello@fa2.ch}{benedetta.franceschiello@fa2.ch}
\and
 Dario Prandi \at
 Universit\'e Paris-Saclay, CNRS, CentraleSup\'elec,  Laboratoire des signaux et syst\`emes, 91190, Gif-sur-Yvette, France.\\
\href{mailto:dario.prandi@centralesupelec.fr}{dario.prandi@centralesupelec.fr}
}

\date{Received: date / Accepted: date}
% The correct dates will be entered by the editor

\maketitle

\begin{abstract}
We consider the evolution model proposed in \cite{Bertalmio2007,BertalmioFrontiers2014} to describe illusory contrast perception phenomena induced by surrounding orientations.  Firstly, we highlight its analogies and differences with the widely used Wilson-Cowan equations \cite{WilsonCowan1973}, mainly in terms of efficient representation properties.  Then, in order to explicitly encode local directional information, we exploit the model of the primary visual cortex (V1)  proposed in \cite{Citti2006} and largely used over the last years for several image processing problems \cite{Duits2010,Prandi2017,Franceschiello2018}.
The resulting model is thus defined in the space of positions and orientation and it is capable to describe assimilation and contrast visual bias at the same time. We report several numerical tests showing the ability of the model to reproduce, in particular, orientation-dependent phenomena such as grating induction and a modified version of the Poggendorff illusion. For this latter example, we empirically show the existence of a set of threshold parameters differentiating from inpainting to perception-type reconstructions and describing long-range connectivity between different  hypercolumns in V1.

% \PACS{PACS code1 \and PACS code2 \and more}
% \subclass{MSC code1 \and MSC code2 \and more}
\end{abstract}

\section{Introduction}  \label{sec:intro}

Recent studies on vision research have shown that many, if not most, popular vision models can be described by a cascade of linear and non-linear (L+NL) operations \cite{Martinez2018}. This is the case for several reference models describing visual perception - e.g. the Oriented  Difference  Of  Gaussians  (ODOG)  \cite{Blakeslee1999} or the Brightness  Induction  Wavelet  Model (BIWaM) \cite{Otazu2008} - and, analogously, for models describing neural activities \cite{Carandini2005}.
These L+NL models are suitable in many cases for describing retinal and thalamic activity, but they have been shown to have low predictive power for modelling the neural activity in the primary visual cortex (V1), explaining less than 40\% of the variance of the data \cite{Carandini2005}. On the other hand, there exist several models in vision research which cannot be expressed as a combination of (L+NL) operations. 
{\color{black}
  Prominent examples are models describing neural dynamics via Wilson-Cowan equations \cite{WilsonCowan1973,Bressloff2001,SartiCitti2015}. Although these models have been extensively studied by the neuroscience community to describe cortical low-level dynamics, see, e.g. \cite{Cowan2016}, their use in the context of psychophysics to describe, e.g., visual illusions has been considered only very recently \cite{JNP2019}.
  }

%{\color{red}Prominent examples are, for instance, models describing neural dynamics via Wilson-Cowan equations \cite{WilsonCowan1973,Bressloff2001}. These equations describe the activation state $a(\xi,t)$ of a population of neurons at time $t>0$ with V1 coordinates $\xi=(x,\theta)$, where $x\in\mathbb R^2$ is the spatial preference  and  $\theta\in \bP^1 \simeq [0,\pi)$ is the orientation preference, via the following ODE:
%\begin{multline}\label{eq:WC1}
%    \frac{\partial}{\partial t} a(\xi,t) = 
%    -\alpha a(\xi,t) \\
%    + \nu \int_{\mathbb R^2\times\bP^1} \omega(\xi \| \xi') \sigma(a(\xi',t))\,d\xi'+h(\xi,t). 
%\end{multline}
%Here, $\alpha,\nu>0$ are fixed parameters, $\omega(\xi\|\xi')$ is a kernel modelling the interaction at two different locations $\xi$ and $\xi'$, $\sigma:\mathbb R\to\mathbb R$ is a non-linear sigmoid saturation function and $h$ represents the external stimulus. Wilson-Cowan models in the form \eqref{eq:WC1} have been extensively studied within the neuroscience community to describe cortical low-level dynamics, see, e.g. \cite{Cowan2016}. However, their use in the context of psychophisics as a tool to describe, for instance, visual illusions has been considered only recently by the authors in \cite{JNP2019}, where a discussion on the lack of a variational counterpart for model \eqref{eq:WC1} is given.}

In \cite{Bertalmio2007,BertalmioCowan2009,BertalmioFrontiers2014} the authors show how  a slight, yet effective, modification of the Wilson-Cowan equation that does not consider orientation admits a variational formulation through an associated energy functional which can be linked to histogram equalisation, visual adaptation and the efficient representation principle, an important school of thought in vision science \cite{Olshausen2000}.
This principle, introduced by Attneave \cite{Attneave1954} and Barlow \cite{Barlow1961}, is based on viewing neural systems through the lens of information theory and states that neural responses aim to overcome neurobiological constraints and to optimise the limited biological resources by self-adapting to the statistics of the images that the individual typically encounters, so that the visual information can be encoded in the most efficient way.
Natural images (and, more generally, images in urban environments) are in fact not random arrays of values, since they present a significant statistical structure. With respect to such statistics, nearby points tend to have similar values; as a result, there is significant correlation among pixels, with a redundancy of $90\%$ or more \cite{Atick1992}, and it would be highly inefficient and detrimental for the visual system to simply encode each pixel independently.
Another very important reason to remove redundant statistical information from the representation is that the statistical rules impose constraints on the image values that are produced, preventing the encoded signal from utilizing the full capacity of the visual channel, which is another inefficient or even wasteful use of biological resources. 
By removing what is redundant or predictable from the statistics of the visual stimulus, the visual system can concentrate on what's actually informative \cite{Rucci2015}.
Remarkably, the efficient representation principle has correctly predicted a number of neural processing aspects and phenomena, and is the only framework able to predict the functional properties of neurons from a very simple principle.
In  \cite{Atick1992}, Atick makes the point that one of the two different types of redundancy or inefficiency in the visual system is the one that happens  if some neural response levels are used more frequently than others: for this type of redundancy, the optimal code is the one that performs histogram equalisation, which can be obtained by means of the modification of the WC model described above.

\paragraph{Contribution} The first contribution of this paper
is to formally prove{\color{black}, in a completely general setting,}  that %{\color{red}in a continuous setting}
Wilson-Cowan equations are non-variational, {\color{black}i.e., they cannot be written as the gradient flow of an $L^2$ energy functional}. For this reason, their solutions do not provide a representation as efficient as the  solutions to the local histogram equalisation model.

\begin{figure*}[t]
    \centering
    \includegraphics[height = 2.5cm]{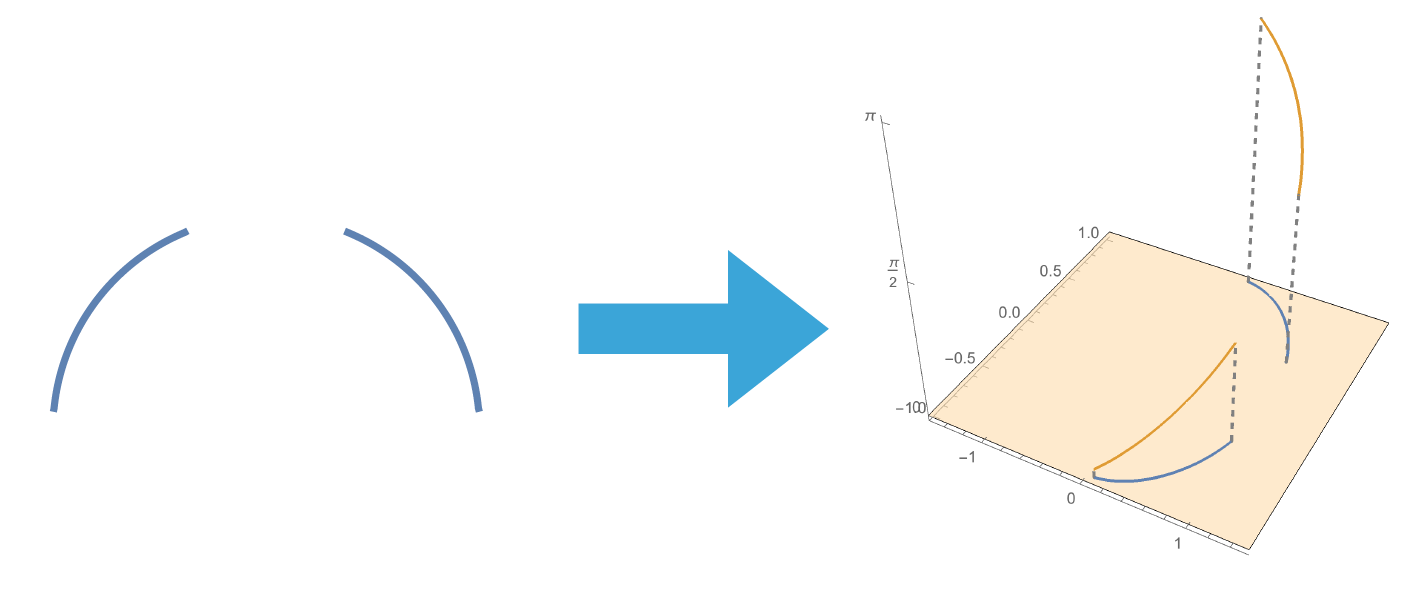}
       \includegraphics[height = 2.5cm]{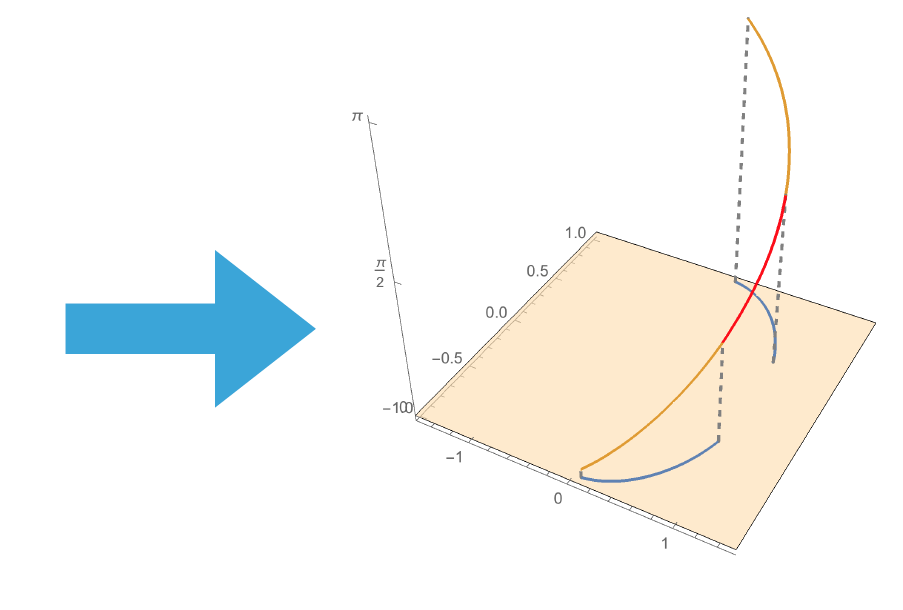} 
        \includegraphics[height = 2.5cm]{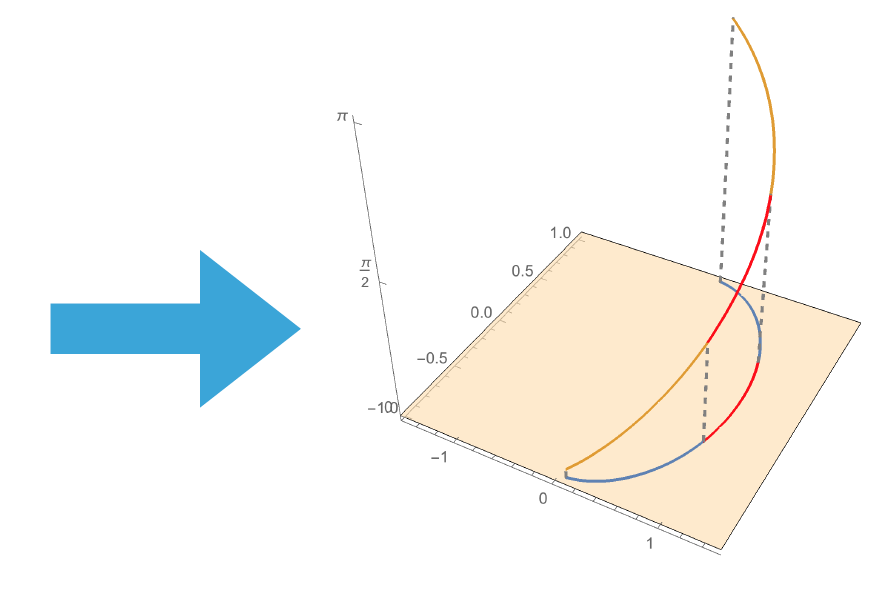}
        \includegraphics[height = 2.5cm]{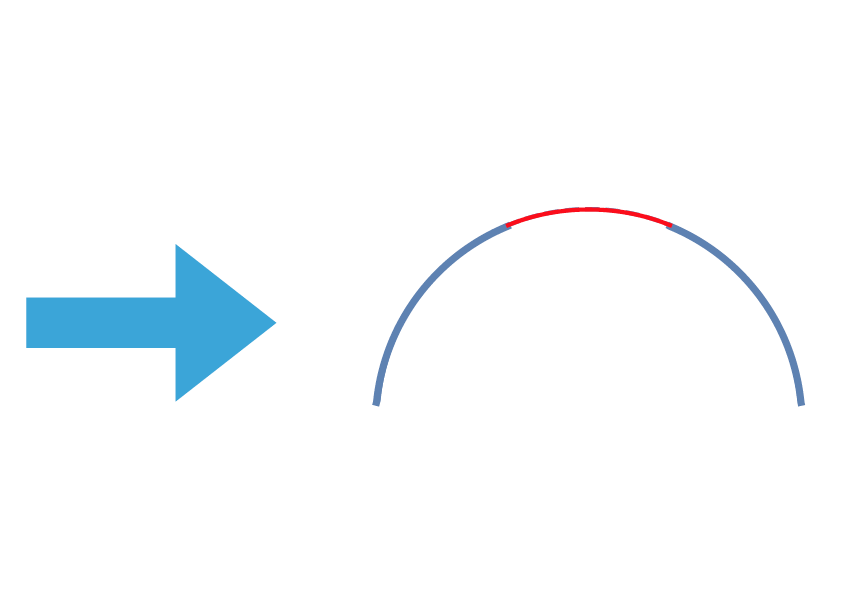}
    \caption{Pipeline for cortical-inspired image processing: Each $x \in \mathbb{R}^2$ is lifted in the space of positions and orientations $\mathbb{R}^2 \times \mathbb{P}^1$ according to the correspondent tangent direction of the curve at point $x$. In the lifted space, many operations can be performed, such as the completion of of the given broken curve. Then, the information retrieved within the lifted space can be projected back to the $\mathbb{R}^2$ plane.}
    \label{fig:lifting}
\end{figure*}

As a secondo contribution, we introduce an explicit orientation dependence %{\color{red}into this}
{\color{black} both into the WC equations and into this} modification via a lifting procedure inspired by the neuro-physiological modelling of V1 \cite{Citti2006,Duits2010,Prandi2017}, which has also been applied to several image processing problems \cite{Boscain2018a,Zhang2016}. The lifting procedure, illustrated in Figure~\ref{fig:lifting}, consists in associating to each point of the retinal plane $x \in \mathbb{R}^2$, the tangent direction $\theta$ of the contour at point $x$,  thus `lifting' the retinal plane $\mathbb{R}^2$ to the feature space $\mathbb{R}^2 \times \mathbb{P}^1$ of positions and orientations. This mathematical construction mimics the neural representations of the image features that the visual cortex performs, as it is well-known from the studies in vision science by Hubel and Wiesel \cite{hubel1968receptive}.

%\textcolor{red}{Following the preliminary version of this work \cite{SSVM2019},} 
We then report some numerical evidence showing how the proposed model is able to better reproduce several visual perception bias than both
its orientation-independent version and  some reference (L+NL) models.
In particular, after reporting some numerical results for classical non-orientation-dependent illusions, we  test our model on orientation-dependent Grating Induction (GI) phenomena (generalising the ones presented in \cite[Figure~3]{Blakeslee1999}, see also \cite{McCourt1982}) and show a direct dependence of the output image on the local orientation, which cannot be described by orientation-independent models.

We then test the proposed model on a modified version of the Poggendorff illusion, a geometrical optical effect where a misalignment of two collinear segments is induced by the presence of a surface \cite{Weintraub1971,Westheimer2008}, see Figure \ref{fig:poggendorff}. For this modified version, our model is able to integrate the contrast feature better than state-of-the-art models  such as those based on filtering techniques, \cite{Blakeslee1999,Otazu2008} on natural images statistics \cite{howe2005poggendorff} and  cortical-based ones \cite{franceschiello2017modelling,franceschiello2019geometrical}. Moreover, we also show that such feature is not correctly integrated by the classical WC equations even when orientation is explicitly taken into account in the modelling.

Finally, %\textcolor{red}{we extend the numerical discussion in \cite{SSVM2019} by further reporting} 
\textcolor{black}{we report} an empirical study concerning the sensitivity of the model to parameters showing the existence of threshold values able to change the nature of the completion properties of the model, e.g. to make it switch from inpainting-type (geometrical completion) to perception-type (perceptual completion).

\textcolor{black}{A preliminary version of this work, including some of the tests presented here, appeared in \cite{SSVM2019}.}

% Wilson-Cowan (WC) equations play a fundamental role in the mathematical description of neural interaction in the first layer of the primary visual cortex (V1). 
% Along with that, they allow to introduce a new class of methods for image processing which go under the name of \emph{Cortical-Inspired}, i.e. where image treatment starts from assumptions on the behaviour of the visual cortex during the act of seeing. 
% Mathematically, this translates into modelling neurons in V1 as filters (such as Cake Wavelets \cite{Bekkers2014}, Gabor Filters \cite{Daugman1985a}, \emph{etc.}) which are able to extract features of the initial image (orientation, scale, frequencies, \emph{etc.}) by means of convolution operations.      
% 
% Recently in \cite{Song2018} a Wilson-Cowan type modelling has further been used to study colour perception phenomena combined with assimilation and contrast description. 
% 
% \paragraph{Contribution.} In this paper we present a cortical-inspired framework for modelling  orientation-induced contrast perception phenomena. Our work is inspired by the recent studies on contrast perception models via Wilson-Cowan-type equations \cite{BertalmioFrontiers2014,KimBatardBertalmio2016} and combined with cortical-inspired lifting procedures \cite{Citti2006,Prandi2017} to codify the redundant information on local direction preferences in an extra variable, via an orientation-dependent filtering of the image  \cite{Bekkers2014}. 
% 

\section{Variational and evolution methods in vision research}  \label{sec:efficient}

The use of variational methods for solving  ill-posed imaging problems is nowadays very classical within the imaging community. 
For a given degraded image $f$ and a (possibly non-linear) degradation operator $\mathcal{T}$ modelling noise, blur and/or under-sampling in the data, the solution of the problem 
\begin{equation} \label{eq:inverse}
    \text{find } u \quad \text{s.t.}\quad f=\mathcal{T}(u)
\end{equation}
often lacks fundamental properties such as existence, uniqueness and stability, requiring alternative strategies to be used in order to reformulate the problem in a well-posed way. 

In the context of variational regularisation approaches, for instance, one looks for an approximation $u_\star$ of the real solution $u$ by solving a suitable optimisation problem, so that
\begin{equation}  \label{eq:prob_var}
  u_\star \in \arg\min \mathcal E(u),
\end{equation}
where $\mathcal E$ is a (possibly non-convex) energy functional which typically combines prior information available both on the image and on the physical nature of the signal (in terms, for instance, of its noise statistics), see, e.g., \cite{ChanShen2005} for a review. 

In convex and smooth scenarios, a common alternative consists in considering the steepest descent of $\mathcal{E}$ defined in terms of the Fr\'echet derivative $\nabla \mathcal{E}$ calculated w.r.t. to some norm, which reduces the problem to the form
\begin{equation}\label{eq:grad_desc}
  \frac{\partial}{\partial t} u = - \nabla\mathcal E(u), \qquad u|_{t=0} = f,
\end{equation}
under appropriate conditions on the boundary of the image domain. %{\color{red}In this formulation,  the solution $u_\star$ in \eqref{eq:prob_var} is found alternatively by looking for} 
Then, solutions $u_\star$ to \eqref{eq:prob_var} correspond to
stationary solutions of \eqref{eq:grad_desc}. 
%
% Historically, evolution equations have been widely studied to model several physical phenomena such as propagation, diffusion and many more. The general formulation \eqref{eq:grad_desc} often offers an insightful understanding on how image information is diffused and/or transported in the object of interest.
% 
 We remark that while the connection between variational problems and parabolic PDEs is always guaranteed by taking the gradient descent of the corresponding energy functional as above, the reverse is not always possible, as it requires some additional  structure of the functional space considered that may lack in several cases. We will comment on this issue in the next section, where we will provide some examples in this respect focusing at some neuro-physiologically inspired models for vision.
 
 In such context, evolution equations have been originally used as a tool to describe the physical transmission, diffusion and interaction phenomena of stimuli in the visual cortex, see, e.g. \cite{Cowan2016}. Similarly, variational methods have been studied by the vision community to describe \emph{efficient neural coding} properties, see, e.g. \cite{Webster2015,Olshausen2000}, i.e. all the mechanisms used by the human visual system to \emph{optimise} the visual experience via the reduction of redundant spatio-temporal biases linked to the perceived stimulus. 
 
 In the context of vision, a first study on the efficient representation aspects of some neuro-physiological model analogous to the one considered in this work, has been recently performed by the authors in \cite{JNP2019} where several visual illusions are studied.

\subsection{Wilson-Cowan-type models for neuronal activation}

A prominent example of evolution models describing neuronal dynamics are the Wilson-Cowan (WC) equations \cite{WilsonCowan1973,Bressloff2001}{\color{black}, that we present here in a general context.}

Consider a neuronal population parametrised by a set $\Omega$, endowed with a measure $d\xi$ supported on the whole $\Omega$. {\color{black} In the following sections we will be interested in the two cases $\Omega = \bR^2$ and $\Omega = \bR^2\times \bP^1$, both endowed with the corresponding Lesbegue measure.}
Denoting by $a(\xi,t)\in\bR$ the state of a population of neurons with coordinates $\xi\in\Omega$ at time $t>0$, the Wilson-Cowan model reads
\begin{multline}
  \label{eq:WC}
    \frac{\partial}{\partial t} a(\xi,t) = -\beta a(\xi,t) \\
    +\nu \int_{\Omega} \omega(\xi \| \xi') \sigma(a(\xi',t))\,d\xi' + h(\xi,t). \tag{WC}
\end{multline}
Here, $\beta>0$ and $\nu\in\R$ are fixed parameters, $\omega(\xi\|\xi')$ \textcolor{black}{is a kernel that models} interactions at two different locations $\xi$ and $\xi'$,  the function $h$ represents an external stimulus  and $\sigma:\R\to \R$ is a non-linear sigmoid saturation function. 

\begin{figure*}[htp]   
    \centering
    \begin{subfigure}{0.45\textwidth}
    \centering
    \includegraphics[width=\textwidth]{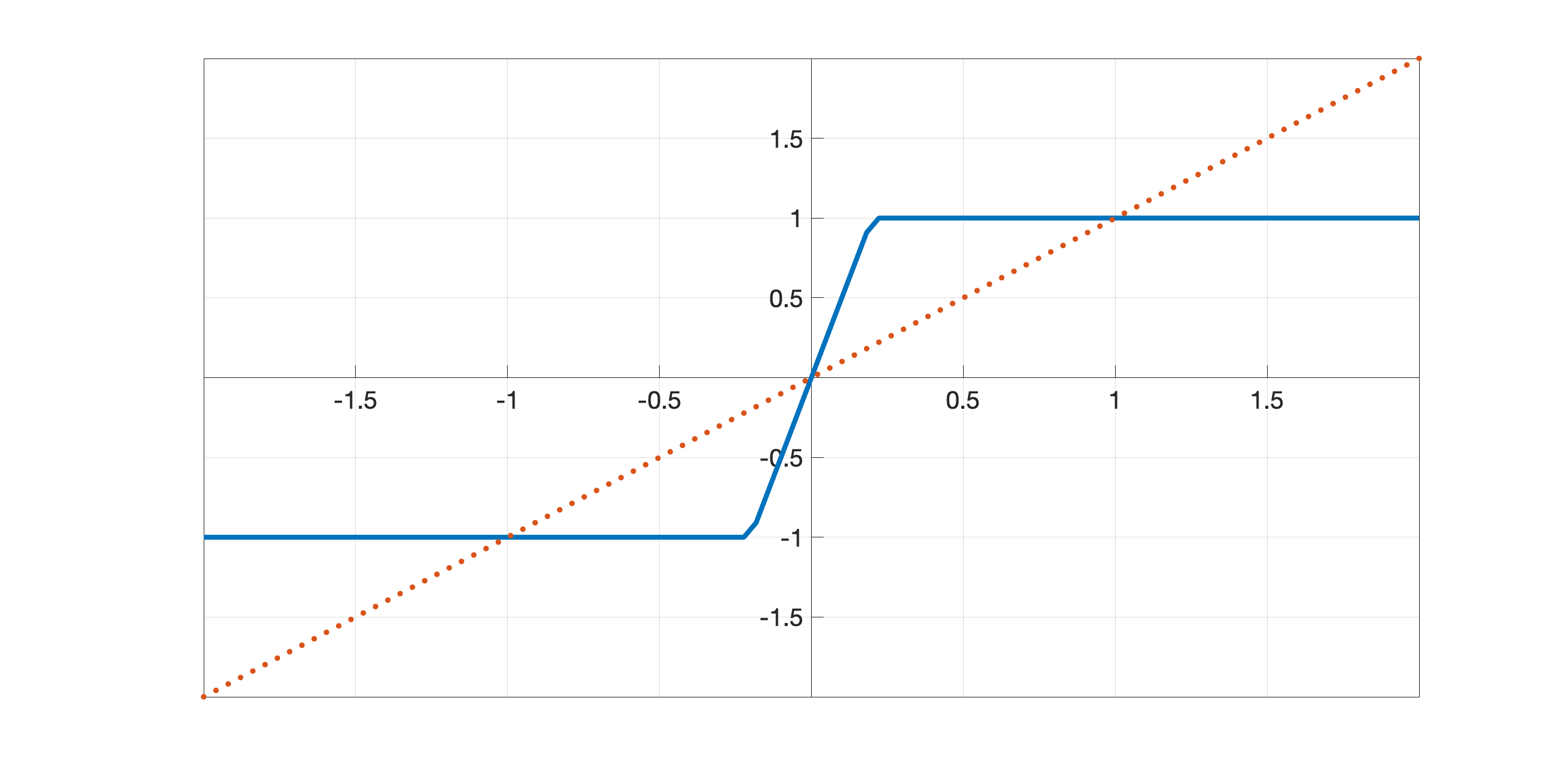}
    \caption{Excitatory sigmoid $\sigma$}
    \end{subfigure}
    \begin{subfigure}{0.45\textwidth}
    \centering
    \includegraphics[width=\textwidth]{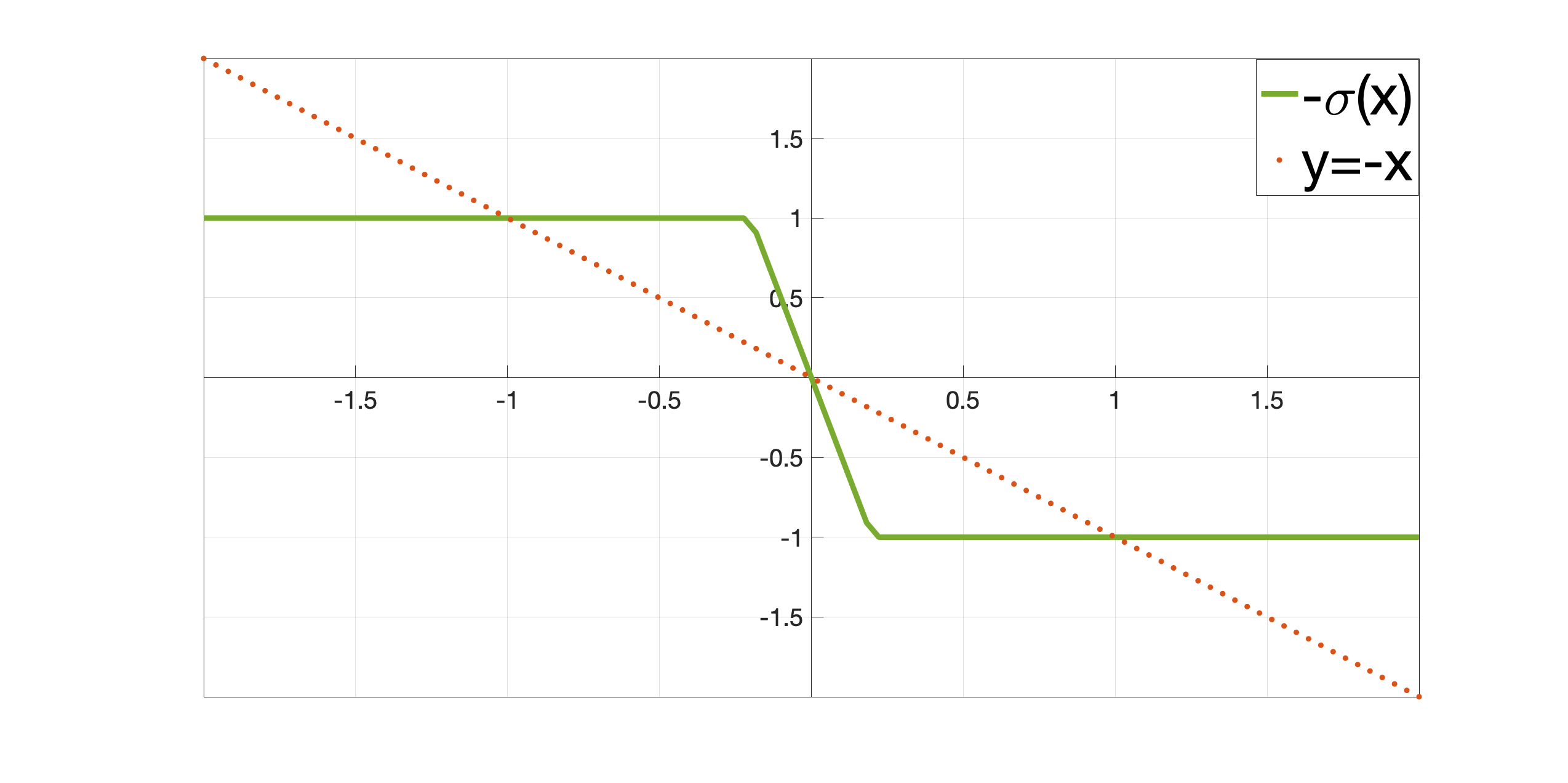}
    \caption{Inhibitory sigmoid $\widehat{\sigma}=-\sigma$}
    \end{subfigure}
    \caption{Symmetric behaviour of excitatory and inhibitory sigmoid functions in the form \eqref{def:sigmoid} with $\alpha=5$. }
      \label{fig:sigmoids}
\end{figure*}

In the following we further assume that the interaction kernel $\omega$ is non-negative and normalised:
\begin{equation}\label{eq:omega-norm}
  \int_{\Omega} \omega(\xi \| \xi')\,d\xi' = 1, \qquad \text{for a.e.\ }\xi\in\Omega.
\end{equation}
Moreover, as a sigmoid $\sigma$, we consider the following odd function:
\begin{equation}  \label{def:sigmoid}
  \sigma(\rho) := \min\{1,\max\{\alpha\rho, -1\}\}, \qquad \alpha>1,
\end{equation}
which has been previously considered, e.g., in \cite{Bertalmio2007}. Observe that, depending on the sign of $\nu$, model \eqref{eq:WC} is able to describe both excitatory ($\nu >0$) and inhibitory local interactions ($\nu<0$), see, e.g. \cite[Section~3]{Bressloff2002}. Due to the oddness of $\sigma$, this latter case can be equivalently expressed by keeping $\nu>0$ and replacing $\sigma$ with its ``mirrored'' version $\hat\sigma(\rho) = \sigma(-\rho), ~\rho\in\R$, see Figure~\ref{fig:sigmoids}.% We plot in Figure \ref{fig:sigmoids} the sigmoids considered in this work in these two cases for $\alpha=5$.

Equation \eqref{eq:WC} has been studied intensively over the last decades to describe several neuronal mechanisms in V1, see, e.g.~, \cite{Faugeras2009,Veltz2009,Cowan2016,Barbierietal2014,SartiCitti2015}. However, one interesting aspect which, up to our knowledge, has not been previously investigated, is whether \eqref{eq:WC} complies with any efficient representation principle, or, in more mathematical terms, whether such model can be interpreted as the gradient descent in the form \eqref{eq:grad_desc} of some energy functional {defined \color{black}on $L^2(\Omega)$}.

As a first result, we show in the following that {\color{black} the model \eqref{eq:WC}} does not satisfy a variational principle. As a consequence, it does not implement an efficient neural coding mechanism. A preliminary study has been performed by the authors in \cite{JNP2019}, in a completely discrete setting. Here, we make these considerations more rigorous by the following theorem.

\begin{theorem}\label{thm}
  Assume that {\color{black} there exist} two subsets of positive measure $U_1,U_2\subset\Omega$, $U_1\cap U_2 = \varnothing$ such that $\omega(\xi\|\xi')> 0$ for any $\xi\in U_1$ and $\xi'\in U_2$. Then, for $\sigma$ chosen as above, the Wilson-Cowan equation \eqref{eq:WC} does not admit a variational formulation, {\color{black}that is, it cannot be expressed as the gradient descent in the Fr\'echet sense of any densely defined energy $\mathcal{E}$. }
\end{theorem}

{\color{black}
\begin{proof}
  We proceed by contradiction and assume that there exists a densely defined energy $\mathcal E$ on $L^2(\Omega)$ such that \eqref{eq:WC} can be expressed in the form \eqref{eq:grad_desc}. 
  
  Let $\chi_i:\Omega\to \{0,1\}$ be the characteristic function of $U_i$, $i=1,2$. Since up to reducing it we can always assume $U_i$ to have finite measure, we have that $\chi_i\in L^2(\Omega)$.   
  Then, we define $J:\bR^2\to\bR$ by 
  \begin{equation}
    J(v) := \cE(v_1\chi_1 + v_2\chi_2), \qquad v=(v_1,v_2)\in\bR^2.
  \end{equation}
  
  By definition, we have 
  \begin{equation}
    \partial_i J(v)=\langle\nabla \cE(v_1\chi_1 + v_2\chi_2), \chi_i\rangle, \qquad i=1,2.
  \end{equation}
  Here, $\nabla \mathcal E$ denotes the Fr\'echet derivative of $\mathcal E$, and $\langle\cdot,\cdot\rangle$ denotes the scalar product in $L^2(\Omega)$.
  Thus, by \eqref{eq:grad_desc} and \eqref{eq:WC}, there holds
  \begin{multline}
    \partial_i J(v) = \beta v_i - \langle h,\chi_i\rangle \\
    - \nu \int_\Omega \int_\Omega \omega(\xi\|\xi')\chi_i(\xi) \sigma\left(\sum_{k=1}^2 v_k \chi_k(\xi')\right)\,d\xi\,d\xi'.
  \end{multline}
  We now aim to differentiate again the above w.r.t.\ the $j$-th variable, $j=1,2$.
  Observe that since $\sigma$ is Lipschitz continuous, it is differentiable almost everywhere and, thanks to the fact that $U_1\cap U_2=\varnothing$, for a.e.\ $v\in \bR^2$ we have
  \begin{equation}
    \sigma'\left(v_1\chi_1(\xi')+v_2\chi_2(\xi')\right) = \sigma'(v_j )\qquad\forall \xi' \in U_j.
  \end{equation}
  This shows that for a.e.\ $v\in \bR^2$ it holds
  \begin{equation}\label{eq:second-diff}
    \partial_{ji} J(v) = \beta\delta_{ij}
    -\nu \sigma'\left( v_j \right) c_{ij},
  \end{equation}
where $\delta_{ij}$ is the Kroenecker delta symbol, and $c_{ij}$ is defined as
  \begin{equation}
    c_{ij} := \int_{U_j\times U_i}\omega(\xi\|\xi')\,d\xi\,d\xi'.
  \end{equation} 
  Observe that, by \eqref{eq:omega-norm}, the assumption on $U_1$ and $U_2$, and the fact that they have finite measure, it holds that $0\le c_{ij}<+\infty$ for any $i,j\in\{1,2\}$. Moreover, since $\omega(\xi\|\xi')>0$ for $\xi\in U_1$ and $\xi'\in U_2$ we have that $c_{21}> 0$.
  
  We now claim that $J\in C^2(A\times A)$, where $A = \{t\in\bR : |t|\neq 1/\alpha\}$ is the set of differentiability of $\sigma$. 
  To this purpose, we compute
  \begin{equation}
    \sigma'(t) = 
    \begin{cases}
      \alpha &  \text{if } |t|<1/\alpha,\\
      0 &  \text{if } |t|>1/\alpha,
    \end{cases}
    \qquad \forall t\in A.
  \end{equation}
  Then, by \eqref{eq:second-diff}, for any $v\in A\times A$ and $i,j\in\{1,2\}$ it holds
    \begin{equation}\label{eq:aaa}
    \partial_{ji} J(v)=
        \begin{cases}
            \beta \delta_{ij} - \nu\alpha c_{ij} & \text{if } |v_j|<1/\alpha,\\
            \beta \delta_{ij}& \text{if } |v_j|>1/\alpha.
        \end{cases}
    \end{equation}
    As a consequence, $\partial_{ji}J$ is continuous on $A\times A$, proving the claim.
  
  To conclude the proof, we show that $\partial_{21}J\not\equiv \partial_{12}J$, which contradicts the $C^2$ differentiability of $J$ by the Schwarz theorem and thus shows that the r.h.s.\ of \eqref{eq:WC} cannot be the Fr\'echet derivative of an energy. Indeed, it suffices to consider $v\in A\times A$ with $v_1>1/\alpha$ and $v_2<1/\alpha$, which by \eqref{eq:aaa} implies
  \begin{equation}
    \partial_{12} J(v) -  \partial_{21} J(v)= \nu\alpha c_{21} \neq 0
  \end{equation}
  This completes the proof of the statement.
\end{proof}
}

\begin{remark}  \label{remark:sigmoid}
  The above argument can be easily extended to any Lipschitz-continuous sigmoid $\sigma$ with non-constant derivative.
\end{remark}

{\color{black}
\begin{remark} The variational nature of physical models describing neural interaction has been investigated in other contexts. For instance, in \cite{Hopfield82}, the authors consider neural models eventually arising from asymmetric interaction kernels. 
We also refer to \cite{French04} for the identification of a Lyapounov functional for a Wilson-Cowan-like equation.
\end{remark}
}

To overcome %\textcolor{red}{this problem} 
\textcolor{black}{the non-existence of an underlying energy for \eqref{eq:WC}} and deal with a model complying with the efficient representation principle, we will consider in the following a variation of \eqref{eq:WC}, which has been introduced in \cite{Bertalmio2007} 
for Local Histogram Equalisation (LHE) {\color{black} of images in the particular case where $\Omega $ is a square domain in $\bR^2$. Keeping now $\Omega$ general and}  using the same notation as above, this model can be written as
\begin{align}\label{eq:LHE}
    \frac{\partial}{\partial t} a(\xi,t) &= -\beta a(\xi,t) \tag{LHE}\\
     &+ \nu \int_{\Omega} \omega(\xi \| \xi') \sigma(a(\xi,t)-a(\xi',t))\,d\xi' + h(\xi,t). \notag
\end{align}

We note that the only difference between \eqref{eq:LHE} and \eqref{eq:WC} is the different input of the sigmoid $\sigma$ appearing inside the integral. While in \eqref{eq:WC} this is equal to the stimulus intensity at location $\xi'$, in \eqref{eq:LHE} this is equal to %{\color{red}a local} 
the difference between the population at the point under consideration and %{\color{red} a neighbouring one}
the neighbouring ones. %Therefore, \eqref{eq:LHE} can be considered as a variant of \eqref{eq:WC} enforcing non-linear behaviour and saturation effects to the local image contrast rather than to the image itself. 

Following the same line of proof as in \cite{Bertalmio2007}, and letting $\Sigma:\R\to \R$ be any (even) primitive function of $\sigma$, it is easy to show  that {\color{black} independently on the choice of $\Omega$,} equation \eqref{eq:LHE} is the gradient descent in the sense of \eqref{eq:grad_desc} of the following energy functional
\begin{multline}  \label{eq:funct_LHE}
  \cE(a) = \frac{\beta-1}{2}\int_{\Omega} \left|a(\xi)\right|^2 \, d\xi +\frac12\int_{\Omega} \left|a(\xi)-h(\xi)\right|^2 \, d\xi \\
  + \frac{\nu}{2}\int_{\Omega}\int_{\Omega} \omega(\xi\| \xi')\Sigma(a(\xi)-a(\xi')) \, d\xi'\,d\xi.
\end{multline}
{\color{black} The functional $\cE$ is the sum of three different terms: The first two can be thought of as data fitting terms whose minimisation forces the solution of \eqref{eq:funct_LHE} to stay close to the given stimulus and, possibly, to global average brightness intensity levels;
the third one is an interaction term whose minimisation corresponds to maximise the local contrast (see \cite{Bertalmio2007} for more details). %{\color{red}The parameters $\beta$ and $\nu$ are typically chosen as $\beta\geq 1$ and $\nu=1/2M$ where $M>0$ is a suitable normalisation constant.} 
In the following section we will make precise some specific choices of $h$, which will clarify the different ingredients of model \eqref{eq:funct_LHE} in more detail.}

\subsubsection{Orientation-independent modelling}

We now focus on the application of \eqref{eq:LHE} to describe contrast perception phenomena {\color{black}independent on local orientation information. To do that, we recall in the following the specific instance of the \eqref{eq:LHE} model introduced in \cite{Bertalmio2007}.}  We model the visual plane as a rectangular domain $Q\subset \bR^2$ and consider grey-scale visual stimuli to be functions $f:Q\to [0,1]$, such that $f(x)$ encodes the brightness intensity at $x$. For a given initial stimulus $f_0$ we then denote by $\mu$ its local {\color{black} average intensity} computed as the convolution $\mu = g\star f_0$ of $f_0$ with some filter $g\in L^1(Q)$ with $\int_Q g(x)\,dx = 1$. 
%\textcolor{red}{Simple Gaussian filters have been considered in \cite{BertalmioFrontiers2014}, whereas a sum of Gaussian filters has been considered in \cite{KimBatardBertalmio2016} to describe multiple inhibition effects happening at a retinal-level \cite{MarceloPLOS2016}. 
%Note that $\mu$ can itself encode a global reference to the Grey-World (GW) principle \cite{Bertalmio2007} by simply setting $\mu(x)\equiv 1/2$ for any $x$.}

\textcolor{black}{In \cite{Bertalmio2007} the filter $g$ was chosen to be uniform, while in \cite{BertalmioFrontiers2014} it was changed to a simple Gaussian in order to reproduce visual induction effects; another possibility for $g$ would be to use a sum of Gaussians, which has been shown to better approximate lateral inhibition effects happening at the retinal-level \cite{MarceloPLOS2016}. }
\textcolor{black}{We also take the activation in  \eqref{eq:LHE} to be $a: = f-1/2$, corresponding to the way our visual system encodes contrast (i.e. as the difference with respect to the average, which we take to be $\frac{1}{2}$). For the external stimulus $h$, we take a weighted sum of the initial stimulus $a|_{t=0} = f_0-1/2$ and its filtering via $g$; in the visual system, this corresponds to a combination of magnocellular (spatially averaged) and parvocellular (fine detail) pathway information.  Namely, for $\lambda>0$, we consider:}
%We further assume that the activation in  \eqref{eq:LHE} is given by $a = f-1/2$ and that the external stimulus $h$ is given by a weighted sum of the initial stimulus $a|_{t=0} = f_0-1/2$ and its filtering via $g$.  Namely, for $\lambda>0$,
\begin{equation}\label{eq:h}
  h = (g\star a)|_{t=0} + \lambda a|_{t=0} = \mu+\lambda f_0 - \frac{1+\lambda}2.
\end{equation}

%\textcolor{red}{Simple Gaussian filters have been considered in \cite{BertalmioFrontiers2014}, whereas a sum of Gaussian filters has been considered in \cite{KimBatardBertalmio2016} to describe multiple inhibition effects happening at a retinal-level \cite{MarceloPLOS2016}. 
%Note that $\mu$ can itself encode a global reference to the Grey-World (GW) principle \cite{Bertalmio2007} by simply setting $\mu(x)\equiv 1/2$ for any $x$.}

%\textcolor{black}{We recall that in \cite{Bertalmio2007} the filter $g$ was chosen to be uniform. It was then  changed to a simple Gaussian in  \cite{BertalmioFrontiers2014}, in order to reproduce visual induction effects; another possibility for $g$ would be to use a sum of Gaussians, which has been shown to better approximate lateral inhibition effects happening at retinal-level \cite{MarceloPLOS2016}. }

{\color{black}We stress that the input $h$  is time-independent. Such simplification follows from considering in our modelling the very short time-frame where the stimulus is presented and retained by the visual system, a time length that is typically known as iconic-memory. For visual illusions such as the ones presented in  Section \ref{sec:numres}, this time-frame typically spans less than 200 ms \cite{Sugita2018VisualPM}, which corresponds to the fixation time between rapid eye movements, and therefore the temporal changes in $h$ can be neglected.}

By plugging the above ingredients in equation \eqref{eq:LHE}, and letting $\beta = 1+\lambda$, we obtain the following (orientation-independent) LHE evolution model:
\begin{align}   \label{eq:LHE2D}\tag{LHE-2D}
%\label{eq:WC2D}\tag{WC-2D}
%  \begin{split}
%  &\frac{\partial}{\partial t} f(x,t)
%  = -(1+\lambda) f(x,t)\\
%  &\,+ \nu\int_{Q} \omega(x,y)\sigma\left(f(y,t)-\frac12 \right)\,dy + \left(\mu(x)+ \lambda f_0(x)\right),  
%  \end{split}\\
  &\frac{\partial}{\partial t} f(x,t)
  = -(1+\lambda) f(x,t)\\
  &+ \nu\int_{Q} \omega(x,y)\sigma\big( f(x,t)-f(y,t) \big)\,dy +\left(\mu(x)+ \lambda f_0(x)\right).   \notag
\end{align}

%\textcolor{red}{We stress that our particular choice of $\beta$ is motivated again by the GW principle. In fact, one can check that this is the only choice guaranteeing that the constant visual stimulus $f_0(x)\equiv 1/2$ is indeed a fixed point for this evolution model.}

\begin{remark}
\textcolor{black}{
Re-arranging the \eqref{eq:LHE2D} equation as
\begin{align}   \label{eq:LHE2Db}
\frac{\partial}{\partial t} f(x,t)
  &= \mu(x) - f(x,t)\\
  &\quad+ \nu\int_{Q} \omega(x,y)\sigma\big( f(x,t)-f(y,t) \big)\,dy \notag\\
  &\quad+ \lambda \left(f_0(x) - f(x,t) \right),   \notag
\end{align}
we can better see the effect of each of its terms: The one multiplied by the parameter $\nu$  enhances local contrast, the one multiplied by $\lambda$ penalises the departure from the original function $f_0$, and the term $\mu(x) - f(x,t)$ pushes the solution towards the local mean. Note that if $\mu(x)$ is the constant value $1/2$, it can be considered as a global average, and the solution is then consistent with the so-called Grey World principle that states that in a sufficiently varied scene the average perceived color is a mid-grey, i.e. a mean value of $\frac{1}{2}$ for each color channel \cite{Bertalmio2007,Bertalmio2014image}.
}
\end{remark}

As far as the interaction kernel $\omega$ is concerned, in \cite{KimBatardBertalmio2016} the authors consider a kernel $\omega$ which is a convex combination of two bi-dimensional Gaussians with different standard deviations. While this variation of the model \eqref{eq:LHE2D} is effective in describing \emph{assimilation} effects, the lack of dependence on the local orientation makes such modelling intrinsically not adapted to explain orientation-induced contrast and colour perception effects such as the ones described in \cite{Otazu2008,Self2014,Blakeslee1999}. Reference models capable to explain these effects are mostly based on oriented Difference of Gaussian linear filtering coupled with some non-linear processing, such as the ODOG and the BIWaM models described in \cite{Blakeslee1999,Blakeslee2016} and  \cite{Otazu2008}, respectively. However, despite their good effectiveness in the reproduction of several visual perception phenomena, these models are not based on any neuronal evolution modelling nor on any efficient representation (i.e. variational) principle.

\subsubsection{Orientation-dependent modelling}

We now focus on orientation-dependent models.
For a given visual stimulus $f$, we let $Lf: Q \times [0,\pi) \to \R$ be the corresponding cortical activation in V1, where $Lf(x,\theta)$ encodes the response of the neuron with spatial preference $x$ and orientation preference $\theta$ to the stimulus $f$. Such activation is obtained  via convolution with the receptive fields of V1 neurons, as explained in Appendix~\ref{a:cortical}, see also \cite{Petitot,Citti2006,Duits2010,Prandi2017}. Then, similarly to above, %{\color{red}we consider} 
{\color{black} the model \eqref{eq:LHE} for a cortical activation $a(x,\theta)$ depending on the local V1 coordinate $\xi=(x,\theta)$ 
%{\color{red}(see Appendix \ref{a:cortical} for more details)}
 is obtained as follows. We define  $F:= a+1/2$ to be the visual stimulus } and take as external stimulus $h = L\mu + \lambda Lf_0 - (1+\lambda)/2$ (compare with \eqref{eq:h}). This, combined with the choice $\beta = 1+\lambda$, yields the equation
\begin{align} 
%\label{eq:WC3D}
% \tag{WC-3D}
%\begin{split}
% & \frac{\partial}{\partial t} F(x,\theta,t)
%  = -(1+\lambda) F(x,\theta,t) 
%  \\&+ \nu\int_0^\pi\int_{Q} \omega(x,\theta\|y,\phi)\sigma\Big(F(y,\phi,t)-\frac12 \Big)\,dy\, d\phi\\
%  &+ \left(L\mu(x,\theta)+ \lambda Lf_0(x,\theta)\right), 
%  \end{split}\\
\label{eq:LHE3D}\tag{LHE-3D}
  &\frac{\partial}{\partial t} F(x,\theta,t)= -(1+ \lambda) F(x,\theta,t) \\
  &+\nu\int_0^\pi\int_{Q} \omega(x,\theta\|y,\phi)\sigma\big(  F(x,\theta,t)-F(y,\phi,t) \big)\,dy\,d\phi \notag \\
   &+\left(L\mu(x,\theta)+ \lambda Lf_0(x,\theta)\right).  \notag
\end{align}
{\color{black}Here the kernel $\omega$ depends both on positions $x,y\in\bR^2$ and orientations $\theta,\phi\in[0,\pi)$. A natural choice for this kernel would be the anisotropic heat kernel naturally associated with the V1 connectivity, as considered in \cite{SartiCitti2015}. However, for numerical reasons,  the  results presented in the following are obtained by considering simply 3D Gaussians. }

We remark once again that the  model above  describes the dynamic behaviour of neuronal activations in the 3D space of positions and orientation. As explained in Appendix~\ref{a:cortical}, once a stationary solution is found, the two-dimensional perceived image can be efficiently found by
\begin{equation}\label{eq:proj}
  f(x) = \frac1\pi\int_0^\pi F(x,\theta)\,d\theta.
\end{equation}

\remark%{Regarding the type of interaction}
In the following we will consider the interaction to be excitatory (i.e., $\nu>0$) for both \eqref{eq:LHE2D} and \eqref{eq:LHE3D} models.
%
%The choice of an inhibitory interaction for WC models is motivated by the following observation: consider $f_0(x)\equiv c$ to be constant. Then, $\mu(x) \equiv c$ and equation \eqref{eq:WC2D} reduces to
%\begin{equation}
%  \frac{\partial}{\partial t} f(t) = \nu \sigma\left(f(t)-\frac12\right), \qquad f_0(x)=c.
%\end{equation}
%This equation has an equilibrium at $1/2$ which is asymptotically stable if $\nu<0$ and unstable if $\nu>0$. Thus, the Grey-World principle suggests the inhibitory regime to be the correct one, since in this case every solution converges to the constant value $1/2$, while in the excitatory regime all solutions with $c\neq 1/2$ diverge.
%
%On the other hand, in \eqref{eq:LHE2D} 
Indeed, the integral term in both models is positive at $x$ if, e.g., $f(x,t)>f(y,t)$. Thus, in order to enhance the contrast between $x$ and its surround we need to have $\nu>0$.

We now discuss on the numerical aspects required to implement model \eqref{eq:LHE3D}.

\subsection{Discretisation via gradient descent}\label{sec:num_impl}

We discretise the initial (square) image $f_0$ as an $N\times N$ matrix. For simplicity, we assume here periodic boundary conditions. We additionally consider $K\in \mathbb N$  orientations, parametrised by 
\begin{equation}
  k \in \{1,\ldots,K\}\mapsto \theta_k := \frac{(k-1)\pi}{K}.
\end{equation}

The discretised lift operator, still denoted by $L$, transforms $N\times N$ matrices into $N\times N\times K$ arrays. Its action on an $N\times N$ matrix $f$ is defined for $n,m\in \{1,\ldots, N\}$ and $k\in \{1,\ldots,K\}$ by
\begin{equation}
  (Lf)_{n,m,k} = \mathcal F^{-1}\left( (\mathcal F f) \odot (R_{\theta_k} \mathcal F\Psi^{\text{cake}}) \right)_{n,m},
  %\quad \forall n,m\in \{1,\ldots, N\}, \, k\in \{1,\ldots,K\},
%\ n,m\in \{1,\ldots, N\}, \, k\in \{1,\ldots,K\},
\end{equation}
where $\odot$ is the Hadamard (i.e., element-wise) product of matrices, $\mathcal F$ denotes the discrete Fourier transform, $R_{\theta_k}$ is the rotation of angle $\theta_k$, and  $\Psi^{\text{cake}}$ is the cake mother wavelet (see Appendix~\ref{a:cortical}).

We let $F^0 = Lf_0$, and $G_0 = L\mu$, where the local {\color{black}average intensity} $\mu$ is given by a Gaussian filtering of $f_0$. The explicit time-discretisation of the gradient descent \eqref{eq:LHE3D} is, for $\Delta t\ll 1$ and $\ell\in\mathbb N$,
{\color{black}
\begin{equation}
    \frac{F^{\ell+1} - F^\ell}{\Delta t}
    = 
    -(1+\lambda)F^\ell + G_0 +\lambda F^0 + \nu \mathcal R_{F^\ell},
\end{equation}
 where $\mathcal R_{F^\ell}$ is the discretisation of the integral term in \eqref{eq:LHE3D}. That is,}
 for a given 3D Gaussian matrix $W$  encoding the weight $\omega$, and an  {\color{black}{$N\times N\times K$}} matrix $F$, we let, for any $n,m\in \{1,\ldots, N\}$ and $k\in\{1,\ldots, K\}$, 
\begin{multline}
    (\mathcal R_{F})_{n,m,k} \\:= \sum_{n',m'=1}^N\sum_{k'=1}^K W_{n-n', m-m', k-k'} \sigma( F_{n,m,k} - F_{n',m',k'}  ).
\end{multline}
We refer to \cite[Section ~IV.A]{Bertalmio2007} for the description of an efficient numerical approach  used to compute the above quantity in the 2D case, that can be translated verbatim to the 3D case under consideration.

After a suitable number of iterations $ \bar \ell$ of the above algorithm (measured by the stopping criterion $\|F^{\ell+1}-F^\ell\|_2/\|F^\ell\|_2\le \tau$, for a fixed tolerance $\tau\ll 1$), the output image is then found via \eqref{eq:proj} as 
\begin{equation}
      \bar f_{n,m} = \sum_{k=1}^K F^{\bar \ell}_{n,m,k}.
\end{equation}

\section{\textcolor{black}{Experiments}}  \label{sec:numres}

In this section we present the results obtained by applying the cortical-inspired model presented in the previous section to some well-known phenomena where contrast perception may be affected by local orientations. 

We compare the results obtained by our orientation-dependent 3D model \eqref{eq:LHE3D} with the corresponding 2D model \eqref{eq:LHE2D} already considered in \cite{KimBatardBertalmio2016,BertalmioFrontiers2014} for histogram equalisation and contrast enhancement. We further compare the performance of these models with two standard reference models based on oriented Gaussian filtering: the ODOG \cite{Blakeslee1999}, and the BIWaM model \cite{Otazu2008}. In the former, the output is computed via a convolution of the input image with oriented difference of Gaussian filters in six orientations and seven spatial frequencies. The filtering outputs within the same orientation are then summed in a non-linear fashion privileging higher frequencies. The BIWaM model is a variation of the ODOG model, the difference being the dependence on the local surround orientation of the contrast sensitivity function\footnote{For our comparisons we used the ODOG and BIWaM codes freely available at \url{https://github.com/TUBvision/betz2015_noise}. }.

\textcolor{black}{\paragraph{Prediction of the perceptual outcome} In this study our objective is to understand the capability of these models to \emph{replicate} the visual illusions under consideration. That is, we are interested in whether the output produced by the models qualitatively agrees with the human perception of the phenomena in some specific and clearly visible region of the image called \emph{target}.  Examples of target are the grey central rectangles of \ref{fig:white} (left).
We stress that our study is purely qualitative; it has to be intended as a proof of concept showing how the discussed models can be effectively used to replicate the perceptual effects according to our notion above. To do so, we use \textit{line profiles}, which qualitatively predict the presence of a perceived illusory phenomena by assessing a change of intensity grey levels in the target of each illusion. We do not address here the match of our numerical outcomes with empirical data since those depend on several further experimental conditions \textcolor{black}{(image  size, luminance of the presented stimulus, duration of the stimulus, etc.)} for which a correspondence with the model parameters is not clear. A dedicated study on experiments motivated by psychophysics, addressing the validation of our models and, possibly, allowing for the creation of ground-truth references for a quantitative assessment is  outside of the scope of this paper.}

\paragraph{Parameters.} All the images considered in the following numerical experiments have size $200 \times 200$ pixels. The lifting procedure to the space of positions and orientations is obtained by discretising $[0,\pi]$ into $K=30$ orientations. {\color{black} The parameter $\nu$ is set $\nu=1/2$.} The relevant cake wavelets are then computed following \cite{Bekkers2014}, setting the frequency band $\texttt{bw}=4$ for all experiments. {\color{black}In \eqref{eq:LHE3D}, we compute the local mean average $\mu$ by a 2D Gaussian filtering with standard deviation $\sigma_\mu$ and the integral term by a 3D Gaussian filtering with standard deviation $\sigma_\omega$.} The gradient descent algorithm stops when the relative stopping criterion defined in Section~\ref{sec:num_impl} is verified with a tolerance $\tau = 10^{-2}$.

\subsection{Non-orientation-dependent examples}

\begin{figure*}[t]
    \centering
    \begin{subfigure}{\textwidth}
    \centering
    \includegraphics[width=.24\textwidth]{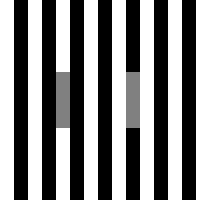}
    \hspace{.05\textwidth}
    \includegraphics[width=.24\textwidth]{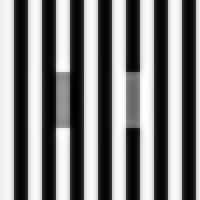}
    \hspace{.05\textwidth}
    \includegraphics[width=.24\textwidth]{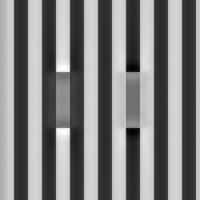}
    \caption{White's illusion.}
    \label{fig:white}
 \end{subfigure}
 \vspace{1em}
 
  \begin{subfigure}{\textwidth}
    \centering
    \includegraphics[width=.24\textwidth]{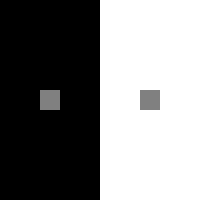}
 \hspace{.05\textwidth}
    \includegraphics[width=.24\textwidth]{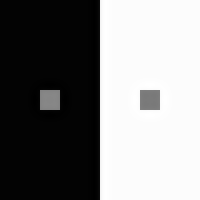}
 \hspace{.05\textwidth}
    \includegraphics[width=.24\textwidth]{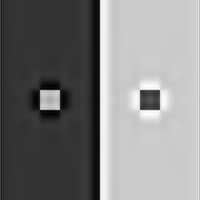}
    \caption{Simultaneous Brightness Contrast illusion.}
     \label{fig:bright}
 \end{subfigure}
 \vspace{1em}
  
 \begin{subfigure}{\textwidth}
    \centering
   \includegraphics[width=.24\textwidth]{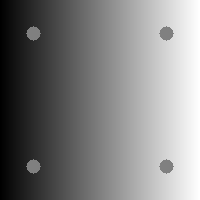}
    \hspace{.05\textwidth}
   \includegraphics[width=.24\textwidth]{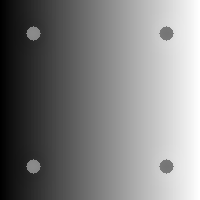}
    \hspace{.05\textwidth}
   \includegraphics[width=.24\textwidth]{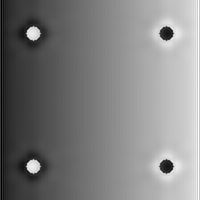}
   \caption{Luminance illusion.}
   \label{fig:lumi}
\end{subfigure}
    \caption{Model output for non-orientation-dependent examples. \emph{First column:} Original image. \emph{Second column:} Output of \eqref{eq:LHE2D} model. \emph{Third column:} Output of \eqref{eq:LHE3D} model.
    Parameters for \eqref{eq:LHE3D}:  $\sigma_\mu=3$, $\sigma_\omega=8$, $\lambda=0.5$.}
    \label{fig:non-or}
\end{figure*}

\begin{figure*}[t]
    \centering
   \begin{subfigure}{.45\textwidth}
    \centering
        \includegraphics[width=.9\textwidth]{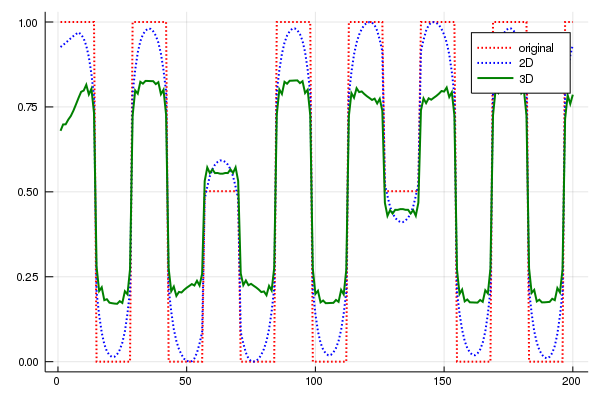}
    \caption{ {\color{black}White's illusion, middle line profile.}}
    \label{fig:pwhite}
    \end{subfigure}
    \begin{subfigure}{.45\textwidth}
    \centering
        \includegraphics[width=.9\textwidth]{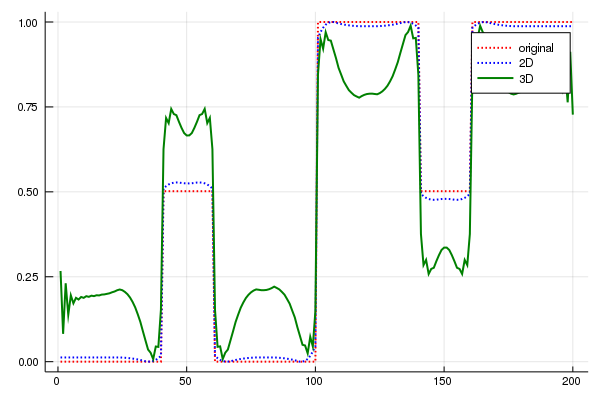}
        \caption{ {\color{black} Brightness illusion, middle line profile.}}
        \label{fig:pbright}
    \end{subfigure}\\
    \begin{subfigure}{.45\textwidth}
    \centering
        \includegraphics[width=.9\textwidth]{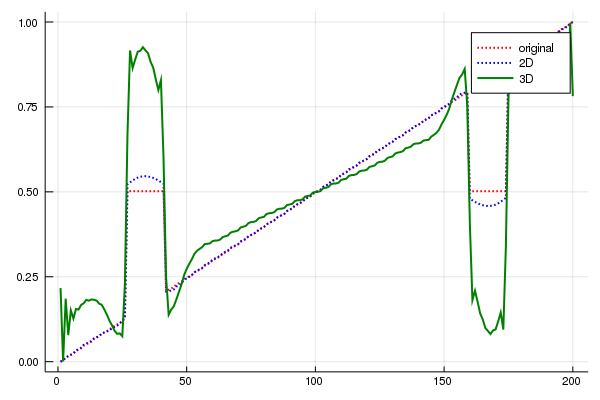}
    \caption{ {\color{black} Luminance illusion, top circles cross line profile.}}
    \label{fig:plumi}
    \end{subfigure}
    \caption{ {\color{black} Line profiles of outputs in Fig.~\ref{fig:non-or}.}}
    \label{fig:pnon-or}
\end{figure*}

In this section we test \eqref{eq:LHE2D} and \eqref{eq:LHE3D} on some classical non-orientation-dependent illusions. 
In particular, we focus on the three following examples:
\begin{enumerate}
\item \emph{White's illusion} \cite{white1979new}, presented in Figure~\ref{fig:white}. Here, the left grey rectangle appears darker than the right one, although their brightness intensity is identical. 
\item The \emph{Simultaneous  Brightness Contrast} illusion \cite{bruke}, presented in Figure~\ref{fig:bright}. It consists in the lighter appearance of the left grey square than the right one. 
\item The \emph{Luminance illusion} \cite{kitaoka} presented in Figure~\ref{fig:lumi}. It consists of  four identical dots over a background whose brightness is smoothly increasing from left to right: the dots on the left are perceived being lighter than the ones on the right. 
\end{enumerate}

We refer the reader to \cite{JNP2019} where more non-orientation-dependent examples are studied.

\paragraph{Discussion.} As Figures~\ref{fig:non-or} \textcolor{black}{and \ref{fig:pnon-or}} show, both \eqref{eq:LHE2D} and \eqref{eq:LHE3D}  predict the three described illusions. \textcolor{black}{Figure~\ref{fig:pnon-or} contains the line profiles relative to the results of Figure~\ref{fig:non-or}. In Figure~\ref{fig:pwhite} we plot the central horizontal line of the images in Figures~\ref{fig:white}, which crosses both grey patches. As the plots show, both models correctly predict the left target to be perceived darker than the right one. Figure~\ref{fig:pbright} contains the plot of the central horizontal line profile of the images in Figure~\ref{fig:bright}, which crosses the two grey squares: both \eqref{eq:LHE2D} and \eqref{eq:LHE3D} correctly predict the left square to be lighter than the right one. Finally, in Figure~\ref{fig:plumi} we plot horizontal profiles crossing top left and right targets (grey circles) of the images in Figure~\ref{fig:lumi}. For each target, both models replicate the brighter perception of the left target with respsect to the right one.}

Notice that  also the BIWaM and ODOG methods can correctly reproduce these illusions (see  \cite{Blakeslee1999,Otazu2008} for numerical results).

\begin{figure*}[t]
\centering
\begin{subfigure}{0.45\textwidth}
    \centering
   \includegraphics[width=.53\textwidth]{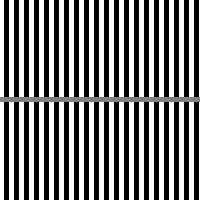}
   \caption{Relative orientation $\theta=\pi/2$.}
   \label{fig:barrette1}
\end{subfigure}
\begin{subfigure}{0.45\textwidth}
\centering
    \includegraphics[width=.53\textwidth]{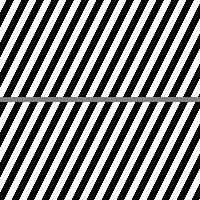}
    \caption{Relative orientation $\theta=\pi/3$.}
     \label{fig:barrette2}
 \end{subfigure}
    \caption{ {\color{black} Grating inductions with different  orientation of the background grating w.r.t. to the central bar.} }
    \label{fig:barrette_mult_or}
\end{figure*}

\begin{figure*}[t]
\centering
\begin{subfigure}{0.24\textwidth}
    \centering
   \includegraphics[width=\textwidth]{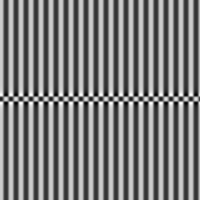}
   \caption{ODOG.}
   \label{fig:recODOG}
\end{subfigure}
 \begin{subfigure}{0.24\textwidth}
\centering
    \includegraphics[width=\textwidth]{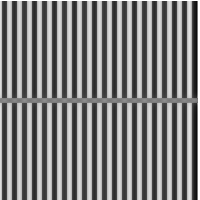}
    \caption{BIWaM.}
     \label{fig:rec3D}
 \end{subfigure}
\begin{subfigure}{0.24\textwidth}
\centering
    \includegraphics[width=\textwidth]{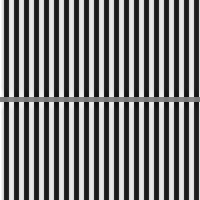}
    \caption{\eqref{eq:LHE2D}.}
    \label{fig:rec2D}
 \end{subfigure}
 \begin{subfigure}{0.24\textwidth}
\centering
    \includegraphics[width=\textwidth]{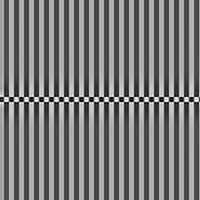}
    \caption{\eqref{eq:LHE3D}.}
     \label{fig:rec3D}
 \end{subfigure}
    \caption{Model outputs for input in Fig.~\ref{fig:barrette1}. \small Parameters for (d): $\sigma_{\mu} = 10$, $\sigma_\omega = 5$, $\lambda = 0.5$.\normalsize}
    \label{fig:reconstructions_barrette}

\centering
\begin{subfigure}{0.45\textwidth}
    \centering
   \includegraphics[width=\textwidth]{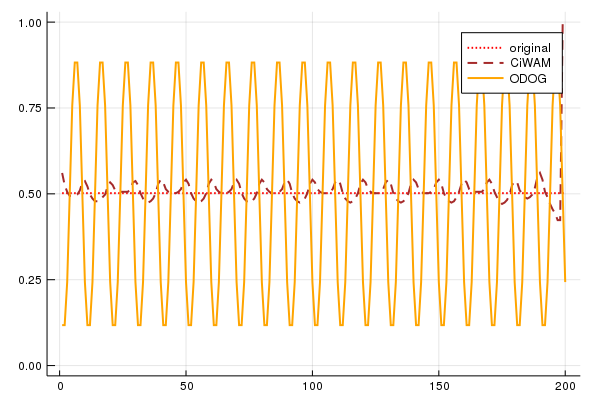}
   \caption{ODOG and BIWaM.}
   \label{fig:lineODOGBIWaM}
\end{subfigure}
\begin{subfigure}{0.45\textwidth}
\centering
    \includegraphics[width=\textwidth]{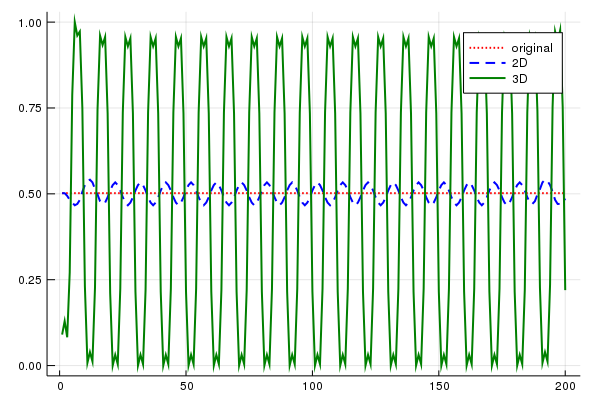}
    \caption{\eqref{eq:LHE2D} VS \eqref{eq:LHE3D} models.}
    \label{fig:line2D3D}
 \end{subfigure}
    \caption{Middle line profiles of outputs in Fig.~\ref{fig:reconstructions_barrette}.  }
    \label{fig:line_profiles_barrette}
\end{figure*}

\subsection{Grating induction with oriented  background}  \label{sec:grating}

Grating induction (GI) is a contrast effect which has been 
first described in \cite{McCourt1982} and later studied, among others, in \cite{Blakeslee1999}. As the name suggests, the phenomenon describes the induction of a regular alternation of intensity changes on a constant image region due to the presence of an inducing background. 

In this section we describe our results on a variation of GI where a relative orientation $\theta$ describes how much the background is oriented with respect to a constant grey bar in the middle of the image, see Figure \ref{fig:barrette_mult_or}.
In such situations, when the background has a different orientation from the central grey bar (i.e. $\theta >0$),  an alternation of dark-grey/light-grey patterns within the central bar is produced and perceived by the observer.
% The effect is stronger when the central grey-bar and the background bars are orthogonal.
This phenomenon is contrast dependent, as the intensity of the induced  grey patterns (dark-grey/light-grey) is in opposition with the background grating.
Moreover, it is also orientation-dependent, since the perceived intensity of the phenomenon varies %increases or decreases 
depending on the background orientation, and, in particular, it is maximal when the background bars are orthogonal to the central one.

\begin{figure*}[t]
\centering
\begin{subfigure}{0.24\textwidth}
    \centering
   \includegraphics[width=\textwidth]{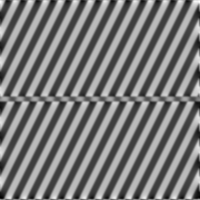}
   \caption{ODOG.}
   \label{fig:recODOG2}
\end{subfigure}
\begin{subfigure}{0.24\textwidth}
\centering
    \includegraphics[width=\textwidth]{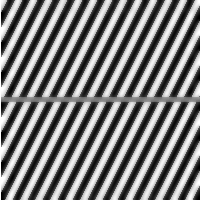}
    \caption{BIWaM.}
     \label{fig:BIWaM2}
 \end{subfigure}
\begin{subfigure}{0.24\textwidth}
\centering
    \includegraphics[width=\textwidth]{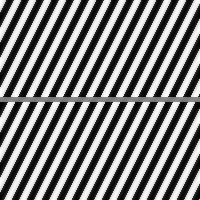}
    \caption{\eqref{eq:LHE2D}.}
    \label{fig:rec2D2}
 \end{subfigure}
 \begin{subfigure}{0.24\textwidth}
\centering
    \includegraphics[width=\textwidth]{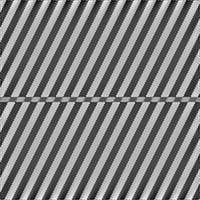}
    \caption{\eqref{eq:LHE3D}.}
     \label{fig:rec3D2}
 \end{subfigure}
    \caption{Model outputs for input in Fig.~\ref{fig:barrette2}. \small Parameters for (d):  $\sigma_{\mu} = 10$, $\sigma_\omega = 5$, $\lambda = 0.5$.\normalsize }
    \label{fig:reconstructions_barrette2}

\centering
\begin{subfigure}{0.45\textwidth}
    \centering
   \includegraphics[width=\textwidth]{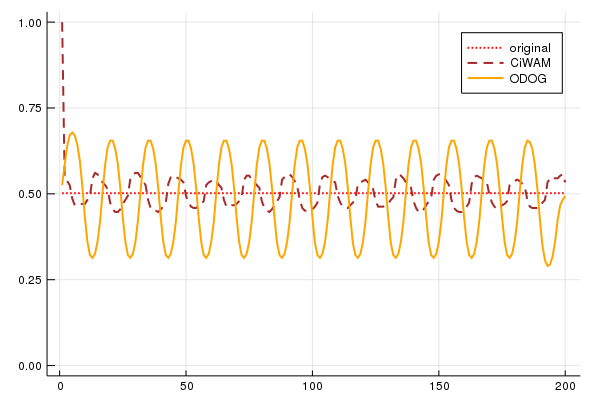}
   \caption{ODOG and BIWaM.}
   \label{fig:lineODOGBIWaM2}
\end{subfigure}
\begin{subfigure}{0.45\textwidth}
\centering
    \includegraphics[width=\textwidth]{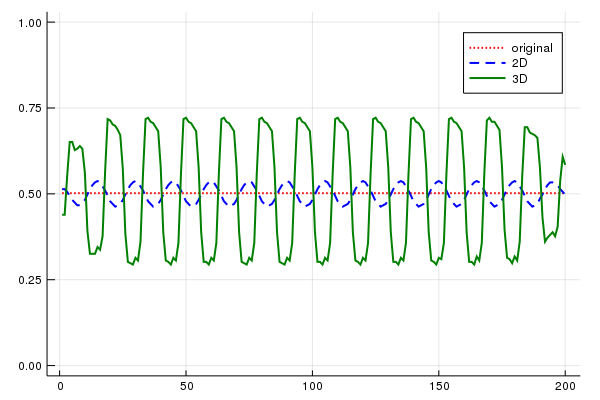}
    \caption{\eqref{eq:LHE2D} VS\eqref{eq:LHE3D} models.}
    \label{fig:line2D3D2}
 \end{subfigure}
    \caption{Middle line profiles of outputs in Figure \ref{fig:reconstructions_barrette2}.  }
    \label{fig:line_profiles_barrette2}
\end{figure*}

\paragraph{Discussion.} We observe that, in accordance with visual perception, model \eqref{eq:LHE3D} predicts the appearance of a counter-phase grating in the central grey bar, see Figures \ref{fig:rec3D} and \ref{fig:rec3D2}. The same result is obtained by the ODOG model, see Figures \ref{fig:recODOG} and \ref{fig:recODOG2}.   Figures \ref{fig:line_profiles_barrette} and  \ref{fig:line_profiles_barrette2} show higher intensity profile when the background gratings are orthogonal to the central line, 
with respect to the case of background angle equal to $\pi/3$,
%while the effect diminishes if the angle of the background decrease from $\pi/2$ to $\pi/3$, 
see orange and green dashed line. On the other hand, BIWaM and \eqref{eq:LHE2D} models do not appear suitable to describe this phenomenon. See for comparison the red and blue dashed lines in Figures \ref{fig:line_profiles_barrette} and  \ref{fig:line_profiles_barrette2}.
% See for comparison the red dashed line in Figure~\ref{fig:lineODOGBIWaM} versus the red dashed line in Figure~\ref{fig:lineODOGBIWaM2}, as well as the blue dashed line in Figure~\ref{fig:line2D3D} versus the one in Figure~\ref{fig:line2D3D2}.

We will now consider a similar example, focusing more precisely on the illusory completion of collinear lines of the background in correspondence of the central grey bar.

\subsection{Poggendorff illusion}  \label{sec:poggendorf}

\begin{figure*}[t]
    \centering
    \begin{subfigure}{0.45\textwidth}
    \centering
    \includegraphics[width=.53\textwidth]{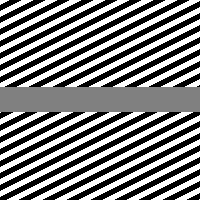}
    \caption{{\color{black}A variation of the Poggendorff illusion with grating. The presence of the grey central surface induces a misalignment of the background lines.}}
    \label{fig:poggendorff-var}
    \end{subfigure}\hfill
    \begin{subfigure}{0.45\textwidth}
       \centering
       \includegraphics[width=.53\textwidth]{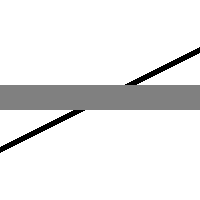} 
   \caption{The classical Poggendorff illusion, extracted from Fig.~ \ref{fig:poggendorff-var}.}
   \label{fig:poggendorff-orig}
   \end{subfigure}
   \\
   \begin{subfigure}{0.45\textwidth}
        \centering
        \includegraphics[width=.53\textwidth]{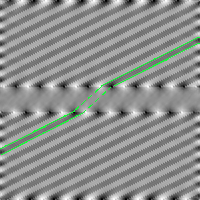}
    \caption{ {\color{black} The perceptual completion computed by using model \eqref{eq:LHE3D}.}}
    \label{fig:poggendorff-out}
    \end{subfigure}
    \hfill
    \begin{subfigure}{0.45\textwidth}
    \centering
       \includegraphics[width=.53\textwidth]{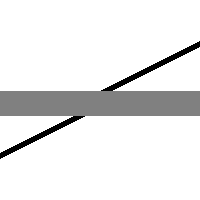}
       \caption{The extracted perceived alignment computed.}
       \label{fig:poggendorff-reconstr}
       \end{subfigure}
    \caption{ {\color{black}Poggendorff illusion: input, detail extraction and result  obtained by \eqref{eq:LHE3D}. Parameters: $\sigma_{\mu} = 3$, $\sigma_\omega = 10$, $\lambda = 0.5$.}}
    \label{fig:poggendorff}
\end{figure*}

The Poggendorff illusion (see Figure~\ref{fig:poggendorff-orig}) consists in the perceived misalignment of two segments of a same continuous line due to the presence of a superposed surface. 
The perceived perceptual bias of the phenomenon has been investigated and studied via neurophysiological experiments, see, e.g., \cite{Weintraub1971,Westheimer2008}. Recently, in \cite{franceschiello2017modelling,franceschiello2019geometrical}, a sub-Riemannian framework where orientations are computed via Gabor filtering has been used to study the geometrical VS. perceptual completion effects induced by the illusion, successfully mimicking human perception. 
Here, we consider a modified version of the Poggendorff illusion, where the background is constituted by a grating pattern, see Figure~\ref{fig:poggendorff-var}, in order to account for both contrast and orientation features. 
%Figure \ref{fig:poggendorff-orig} contains the classical Poggendorff illusion, extracted from figure \ref{fig:poggendorff-var}.

Note that this example is actually similar to the one considered in the previous section, the only difference being the width of the central grey bar, which is the responsible of the perceived misalignment.

\begin{figure*}[t]
\centering
\begin{subfigure}{0.24\textwidth}
    \centering
   \includegraphics[width=\textwidth]{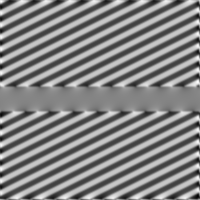}
   \caption{ODOG.}
   \label{fig:POGG_ODOG}
\end{subfigure}
\begin{subfigure}{0.24\textwidth}
\centering
    \includegraphics[width=\textwidth]{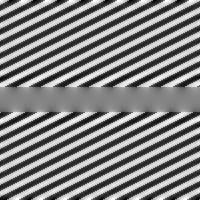}
    \caption{BIWaM.}
    \label{fig:POGG_BIWaM}
 \end{subfigure}
 \begin{subfigure}{0.24\textwidth}
\centering
    \includegraphics[width=\textwidth]{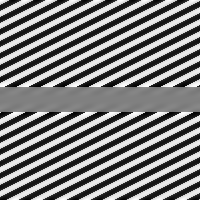}
    \caption{\eqref{eq:LHE2D}.}
    \label{fig:POGG_2D}
 \end{subfigure}
 \begin{subfigure}{0.24\textwidth}
\centering
    \includegraphics[width=\textwidth]{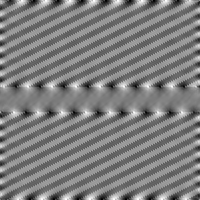}
    \caption{\eqref{eq:LHE3D}.}
    \label{fig:POGG_3D}
 \end{subfigure}
    \caption{{\color{black} Model outputs for the Poggendorff illusion in Fig. \ref{fig:poggendorff-var} via reference models, \eqref{eq:LHE2D}, and \eqref{eq:LHE3D}.}}
    \label{fig:POGG_REF}

\centering
\begin{subfigure}{0.45\textwidth}
    \centering
   \includegraphics[width=\textwidth]{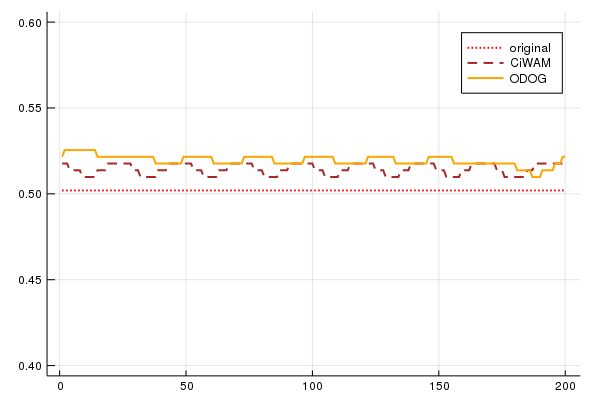}
   \caption{ODOG and BIWaM.}
   \label{fig:lineODOGBIWaM2}
\end{subfigure}
\begin{subfigure}{0.45\textwidth}
\centering
    \includegraphics[width=\textwidth]{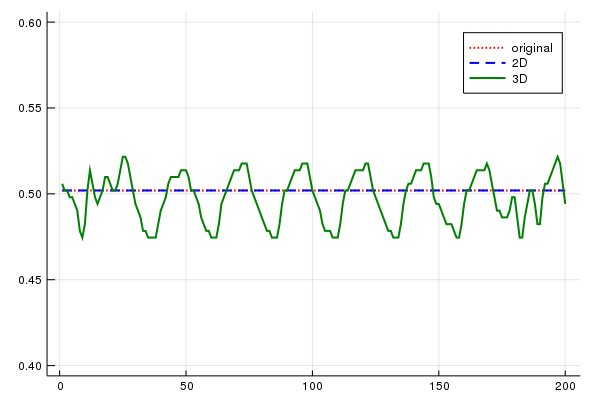}
    \caption{\eqref{eq:LHE2D} VS \eqref{eq:LHE3D} models.}
    \label{fig:line2D3D2}
 \end{subfigure}
    \caption{Middle line profiles of outputs in Figure~\ref{fig:POGG_REF}.  }
    \label{fig:POGG-profiles}
\end{figure*}

\begin{figure*}[t]
    \centering
    \begin{subfigure}{0.24\textwidth}
    \centering
    \includegraphics[width = \textwidth]{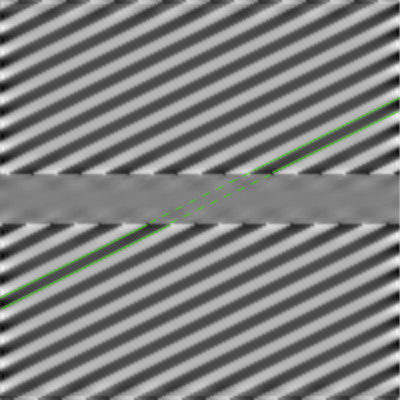}
    \caption{$\sigma_{\omega} = 5$}
    \label{fig:poggendorff-inp1}
    \end{subfigure}\hfill
    \begin{subfigure}{0.24\textwidth}
       \centering
       \includegraphics[width = \textwidth]{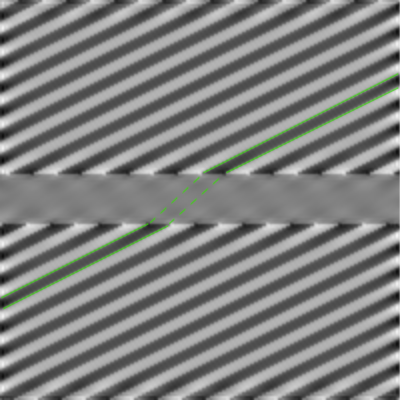} 
   \caption{$\sigma_{\omega} = 6$}
   \label{fig:poggendorff-inp2}
   \end{subfigure}
   \begin{subfigure}{0.24\textwidth}
        \centering
        \includegraphics[width = \textwidth]{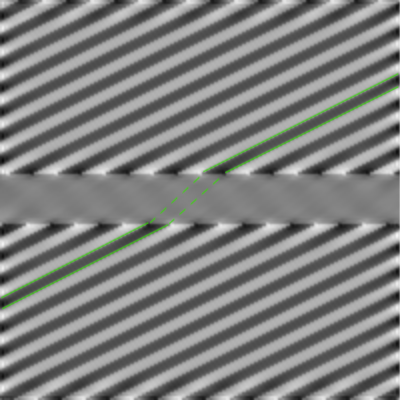}
    \caption{$\sigma_{\omega} = 7$}
    \label{fig:poggendorff-comp1}
    \end{subfigure}
    \hfill
    \begin{subfigure}{0.24\textwidth}
    \centering
       \includegraphics[width = \textwidth]{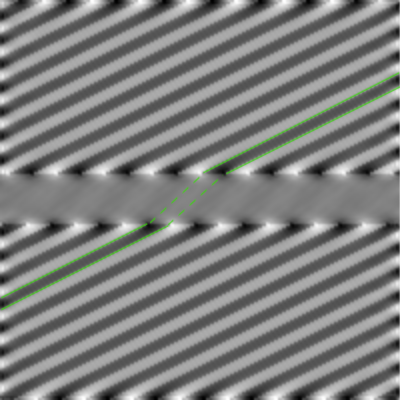}
       \caption{$\sigma_{\omega} = 10$}
       \label{fig:poggendorff-comp2}
       \end{subfigure}
    \caption{Sensitivity to parameter $\sigma_w$ for \eqref{eq:LHE3D} model. The completion inside the middle grey bar changes from geometrical (inpainting-type) to illusory (perception-type). \textcolor{black}{The transition can be observed when varying $\sigma_w = 5$ to $\sigma_w = 6$, while the illusory phenomenon holds for $\sigma_w$ bigger than 6.} \textcolor{black}{The other}  parameters \textcolor{black}{are fixed across experiments}: $\sigma_\mu = 2$, $\lambda = 0.8$.}
    \label{fig:poggendorff}
\end{figure*}

\paragraph{Discussion.} The result obtained by applying \eqref{eq:LHE3D} to Figure~\ref{fig:poggendorff-var} is presented in Figures~\ref{fig:poggendorff-out} and \ref{fig:POGG_3D}. As for the results on the grating induction presented in Section \ref{sec:grating}, we observe an induced counter-phase grating in the central grey bar.

However, the objective of this experiment goes further, the question being whether it is possible to compute numerically an image output reproducing the perceived misalignment between some fixed black stripe in the bottom part of Figure~\ref{fig:poggendorff-var} and its collinear prosecution in the upper part. Note that the perceived alignment differs from the actual geometrical one: for a fixed black stripe in the bottom part, the alignment of the corresponding collinear top stripe is in fact perceived slightly flushed left, see Figure~\ref{fig:poggendorff-orig},  where single stripes have been isolated for better visualisation. The problem here is therefore not an inpainting problem, which is classical in the imaging community, but, rather, to reconstruct the perceptual output from the given input in Fig.~\ref{fig:poggendorff-var}.

We now look at the results in Figure~\ref{fig:poggendorff-out} and mark by a continuous green line a fixed black stripe in the bottom part of the image. In order to find the corresponding perceived collinear stripe in the upper part, we follow how the model propagates the marked stripe across the central surface (dashed green line). We notice that the prosecution computed via the  \eqref{eq:LHE3D} model does not correspond to its actual collinear prosecution, %This can be clearly seen in Figure~\ref{fig:poggendorff-reconstr} where the two stripes have been isolated for better visualisation. 
%This example shows that the proposed algorithm computes an output in agreement with our perception.  
but, rather, it is in agreement with our perception.  
Comparisons with reference models are presented in Figure \ref{fig:POGG_REF} and {\color{black} the corresponding middle-line profiles are shown in Figure~\ref{fig:POGG-profiles}}.
 We observe that the results obtained via the proposed \eqref{eq:LHE3D} model cannot be reproduced by the BIWaM nor the \eqref{eq:LHE2D} models, which moreover induce a non-counter-phase grating in the central grey bar which is different from the expected perceptual result. 
 On the other hand, the result obtained by the ODOG model is consistent with ours, but presents a much less evident alternating grating within the central grey bar. In particular, the induced oblique bands are not visibly connected across the whole grey bar, i.e. their induced contrast is very poor and, consequently, the induced edges are not as sharp as the ones reconstructed via our model, see Figure~\ref{fig:POGG-profiles} forthe middle-line profile.
 
 \begin{figure}[h!]
    \centering
    \includegraphics[width=.24\textwidth]{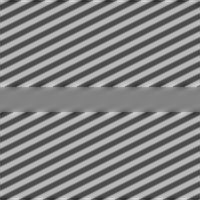}
    \caption{Model output of the standard \eqref{eq:WC} model for the input in Fig.~\ref{fig:poggendorff-var}. See \cite{JNP2019} for other results via the \eqref{eq:WC} models and details on the implementation.}
    \label{fig:poggendorff-WC3D}
\end{figure}
 
 We further remark that a numerical implementation of the standard \eqref{eq:WC} model, whose result is presented in  Figure~\ref{fig:poggendorff-WC3D}, is not able to reproduce the desired perceptual completion.  \textcolor{black}{The model \eqref{eq:LHE3D} reproduces the visual illusion presented in this example better than \eqref{eq:WC}: this is consistent with the variational nature of the model discussed before.} %\textcolor{red}{We believe such failure in replicating the illusion to be  due to the lack of a variational efficient representation as shown in Theorem~\ref{thm}.}

\paragraph{Threshold for inpainting VS. perceptual completion in Poggendorff grating.}
{\color{black}Interestingly, the capability of model \eqref{eq:LHE3D} to reproduce the visual perception bias on the Poggendorff grating example is very much dependent on the choice of the parameter $\sigma_{\omega}$ which accounts for the width of the interaction kernel.

% , which accounts for the width of both these interactions between simple cells in the visual cortex, see Figure \ref{fig:short_long_connect}
As pointed out by the seminal works of Hubel, Wiesel and Bosking \cite{hubel1968receptive,ts1986relationships,bosking1997orientation}, it is possible  to identify at least two main types of connectivity in the visual cortex: the intra-cortical connectivity, able to select the preferred orientations among cells belonging to the same hypercolumn and the long-range connectivity, connecting simple cells belonging to different hypercolumns. 
} 
\begin{figure}[t]
\centering
    \includegraphics[width =0.4\textwidth]{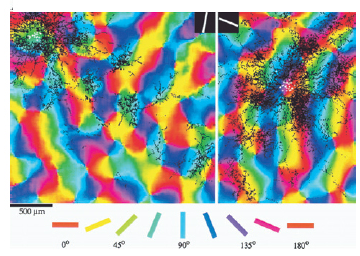}
    \caption{\textcolor{black}{Examples of  long-range (left) and intra-cortical (right) connectivity, from \cite{bosking1997orientation}. Intra-cortical connectivity connects isotropically neurons belonging to the same hypercolumn, while the long range connects those belonging to different hypercolumns, but sensitive to the same orientation.}} 
    \label{fig:short_long_connect}
\end{figure}
Perceptual phenomena such as those presented in this work arise  {\color{black} due to both} these two connectivities, modelled in \eqref{eq:LHE3D} by the parameter $\sigma_{\omega}$ (i.e., the standard variation of the Gaussian $\omega$), therefore accounting for smaller or bigger local interactions.
This parameter can thus be modulated to vary the 
%\textcolor{red}{type of interactions}
\textcolor{black}{width of the connectivity} between different hypercolumns so that when $\sigma_\omega$ is small with respect to the overall size of the processed image, the geometrical completion (inpainting) is reproduced. On the other hand, when $\sigma_\omega$ is large, perceptual-oriented phenomena such as illusory contours or geometrical optical illusions can be modelled. The change between these two types of interactions 
%{\color{red} depending on the size of $\sigma_\omega$}
observed as the parameter $\sigma_\omega$ grows is shown in Figure \ref{fig:poggendorff}. 

This example highlights also the flexibility of our models to adapt to image processing problems and, at the same time but for different choices of parameters, to the modelling of the neural activity in V1.

\section{Conclusions}

In this paper, we considered a neuro-physiological evolution model to study visual perception bias due to contrast and, possibly, to local orientation dependence.  The proposed model has been originally introduced in \cite{Bertalmio2007} in the context of image processing for local histogram equalisation (LHE) and it is a variation of the celebrated Wilson and Cowan equations, formulated in \cite{WilsonCowan1973} to describe the evolution of a population of neurons in V1. 

Firstly, in Section \ref{sec:efficient} we investigated on the efficient representation properties of the original WC model. In mathematical terms this consists in interpreting the corresponding dynamics as the gradient descent of suitable energy functionals. We rigorously prove that for the WC model there is no energy minimised by the WC-dynamics (Theorem~\ref{thm}), while an energy functional  minimised by the stationary solutions of the LHE variant exists (see formula \eqref{eq:funct_LHE}). 

Secondly, by mimicking the structure of V1, we extended the mathematical formulation of the LHE model to a third dimension in order to describe local orientation preferences. This new model, denoted by \eqref{eq:LHE3D}, can be efficiently implemented via convolution with appropriate kernels and solved numerically via standard explicit schemes. The information on the local orientation allows to describe contrast phenomena as well as orientation-dependent illusions.

In Section \ref{sec:numres} we tested this extension of LHE on some orientation-independent brightness illusions, showing that it is able to reproduce the perceptual results as well as standard Linear + Non-Linear filtering (such as the ODOG and the BIWaM models  \cite{Blakeslee1999,Otazu2008}) can do. Then, we performed some further test on orientation-dependent illusions (such as grating induction and the Poggendorff illusion), observing that only the proposed orientation-dependent extension of the LHE model is capable to replicate the perceived visual bias. In agreement with the theoretical sub-optimality of the standard WC model  with respect to the efficient representation principle pointed out before, it turns out that, {\color{black} among the neural-field models tested, the \eqref{eq:LHE3D} model is the one capable of replicating the bigger number of illusions.}

Finally, we reported a preliminary empirical discussion on the sensitivity of model \eqref{eq:LHE3D} to parameters describing different connectivity properties between hypercolumns in V1. Our experiment revealed the existence of a threshold parameter in correspondence of which the completion properties of model \eqref{eq:LHE3D} switch from inpainting-type to perceptual-type. A more accurate theoretical study based, e.g., on bifurcation and stability analysis of the equilibria of the model, is left for future research.

\begin{figure*}[t]
    \centering
    \begin{subfigure}{.3\textwidth}
    \centering
    \includegraphics[width=.8\textwidth]{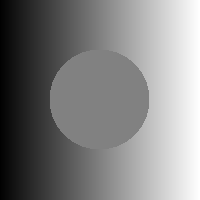}
    \caption{Image 1(A) in \cite{ShapleyGordon85}}
    \label{fig:shapleyA_or}
 \end{subfigure}
 \begin{subfigure}{.3\textwidth}
 \centering
    \includegraphics[width=.8\textwidth]{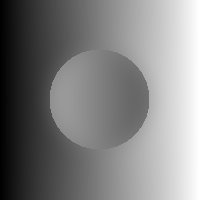}
    \caption{\eqref{eq:LHE2D} reconstruction}
    \label{fig:shapleyA_2d}
 \end{subfigure}
 \begin{subfigure}{.3\textwidth}
 \centering
    \includegraphics[width=.8\textwidth]{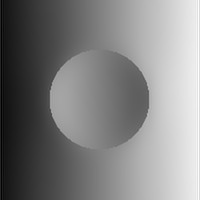}
\caption{\eqref{eq:LHE3D} reconstruction}
    \label{fig:shapleyA_3d} 
    \end{subfigure}
 \vspace{1em}
 
 \begin{subfigure}{.3\textwidth}
    \centering
    \includegraphics[width=.8\textwidth]{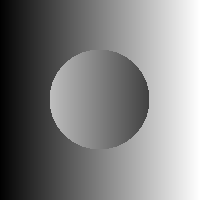}
    \caption{Image 1(B) in \cite{ShapleyGordon85}}
    \label{fig:shapleyB_or}
 \end{subfigure}
 \begin{subfigure}{.3\textwidth}
 \centering
    \includegraphics[width=.8\textwidth]{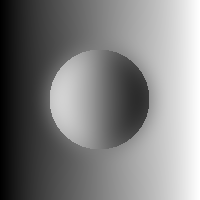}
    \caption{\eqref{eq:LHE2D} reconstruction}
    \label{fig:shapleyB_2d}
 \end{subfigure}
 \begin{subfigure}{.3\textwidth}
 \centering
    \includegraphics[width=.8\textwidth]{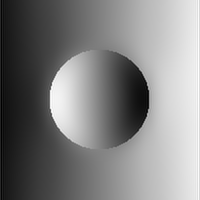}
\caption{\eqref{eq:LHE3D} reconstruction}
    \label{fig:shapleyB_3d} 
    \end{subfigure}
    
    \caption{\textcolor{black}{Reconstruction of Shapley and Gordon illusions, see  \cite[Figures~1(A), 1(B)]{ShapleyGordon85}. \emph{First row:} Image 1(A) in \cite{ShapleyGordon85}. From left to right: original image, reconstruction via the (LHE2D) model, reconstruction via the (LHE3D) model.
    \emph{Second row:} Image 1(B) in \cite{ShapleyGordon85}. From left to right: original image, reconstruction via the (LHE2D) model, reconstruction via the (LHE3D) model.
    Parameters for (LHE3D):  $\sigma_\mu=10$, $\sigma_\omega=50$, $\lambda=0.5$.}}
    \label{fig:shapley}
\end{figure*}

\begin{figure*}[t]
    \centering
    \begin{subfigure}{.45\textwidth}
        \centering
    \includegraphics[width=.7\textwidth]{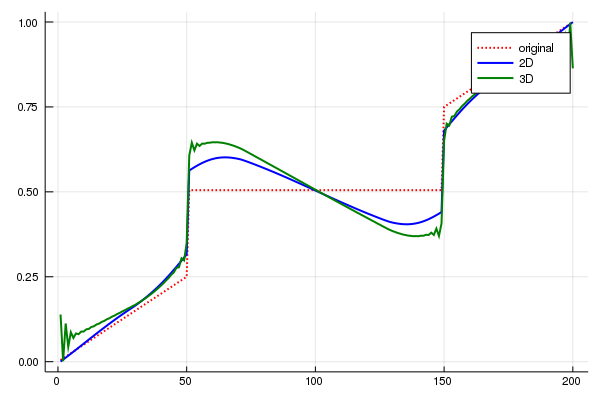}
\caption{Figures \ref{fig:shapleyA_or}, \ref{fig:shapleyA_2d}, \ref{fig:shapleyA_3d}: Horizontal middle line profile.}
\label{fig:pshapleyAh}
    \end{subfigure}
    \begin{subfigure}{.45\textwidth}
        \centering
    \includegraphics[width=.7\textwidth]{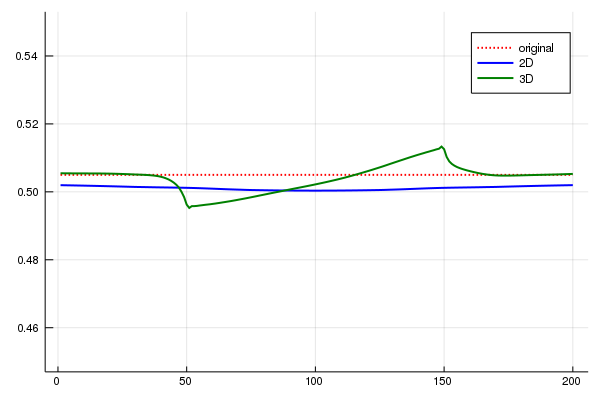}
    \caption{Figures \ref{fig:shapleyA_or}, \ref{fig:shapleyA_2d}, \ref{fig:shapleyA_3d}: Vertical middle line profile.}    
    \label{fig:pshapleyAv}
    \end{subfigure}
    \vspace{1em} 

  \begin{subfigure}{.45\textwidth}
    \centering
    \includegraphics[width=.7\textwidth]{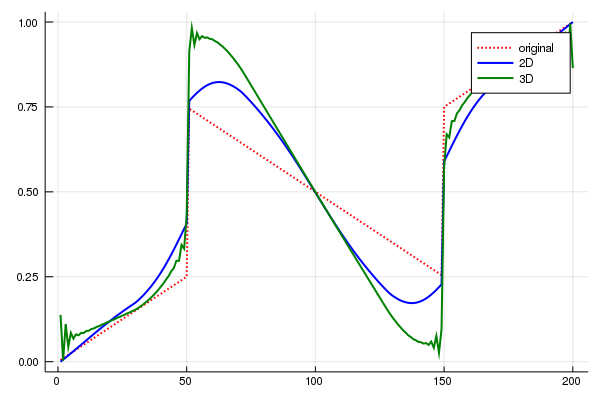}
    \caption{Figures \ref{fig:shapleyB_or}, \ref{fig:shapleyB_2d}, \ref{fig:shapleyB_3d}: Horizontal middle line profile.}
    \label{fig:pshapleyBh}
  \end{subfigure}
    \begin{subfigure}{.45\textwidth}
    \centering 
    \includegraphics[width=.7\textwidth]{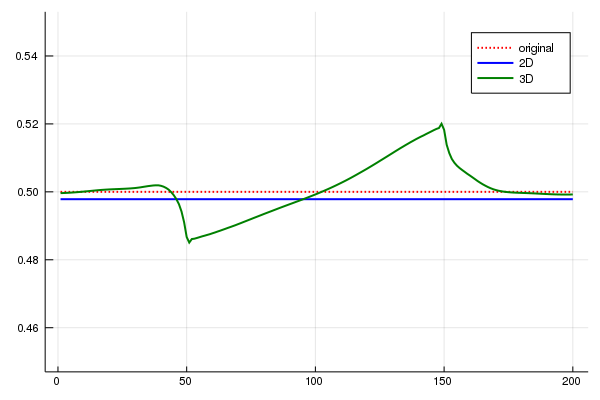}
    \caption{Figures \ref{fig:shapleyB_or}, \ref{fig:shapleyB_2d}, \ref{fig:shapleyB_3d}: Vertical middle line profile.}    \label{fig:pshapleyBv}
    \end{subfigure}
 
    \caption{\textcolor{black}{Line profiles of the images in Figure~\ref{fig:shapley}.}}
    \label{fig:pshapley}
\end{figure*}

Further investigations should also address a more accurate modelling reflecting the actual structure of V1. In particular, this concerns the lift operation where the cake wavelet filters {\color{black} can} be replaced by Gabor filters as in \cite{franceschiello2019geometrical}, as well as the interaction weight $\omega$ which could be taken to be the anisotropic heat kernel of \cite{Citti2006}. % instead of the isotropic Gaussian currently employed.
\textcolor{black}{Furthermore, more features of the image (e.g. scale, frequency, color, etc) should also be considered in future work. According to a preliminary analysis that we performed in this direction, \eqref{eq:LHE3D} seems to be promising when it comes to account for scale. In Figure~\ref{fig:shapleyA_or} we present  a variant of the luminance illusion where only one foreground round patch appears in the image at a bigger scale, while in Figure~\ref{fig:shapleyB_or} we present a variant where the target brightness has a gradient counterposed with respect to the background. The results obtained by applying \eqref{eq:LHE2D} and \eqref{eq:LHE3D} to Figures~\ref{fig:shapleyA_or} and \ref{fig:shapleyB_or} are then reported. We observe that both models correctly reproduce a change of sign in the contrast of the foreground patch, enhancing a 3D effect of the central grey patch. Moreover, \eqref{eq:LHE3D} seems to correctly predict the appearance of an illusory contour in the boundary of the central round patch, as showcased by the line profiles in Figures~\ref{fig:pshapleyAv} and \ref{fig:pshapleyBv}.
  %{\color{red} In fact, in both patterns, the foreground and background brightness are nearly equal in the middle vertical line (see red line profiles in Figures~\ref{fig:pshapleyAv} and \ref{fig:pshapleyBv}), but an illusory contour is perceived, completing the circle. 
  %According to the experiments in \cite{ShapleyGordon85}, the central object has a shaded three-dimensional appearance.}
  %}
}

Extensive numerical experiments should also be performed to assess the compatibility of the model outputs with the psycho-physical tests measuring the perceptual bias induced by these and other phenomena such as the ones discussed in \cite{JNP2019}. This would provide insights about the robustness of the model in reproducing the visual pathway behaviour. 

{\color{black} \begin{acknowledgements}
The authors acknowledge the anonymous referees for their suggestions which improved significantly the quality of their manuscript. 
M.~B. acknowledges the support of the European Union’s Horizon 2020 research and innovation programme under grant agreement number 761544 (project HDR4EU) and under grant agreement number 780470 (project SAUCE), and of the Spanish government and FEDER Fund, grant ref. PGC2018-099651-B-I00 (MCIU/AEI/FEDER, UE).
L.~C., V.~F.\ and D.~P.\ acknowledge the support of a public grant overseen by the French National Research Agency (ANR) as part of the \emph{Investissement d'avenir program}, through the iCODE project funded by the IDEX Paris-Saclay, ANR-11-IDEX-0003-02 and of the research project \emph{LiftME} funded by INS2I, CNRS. V.~F.\ acknowledges the support received from the European Union's Horizon 2020 research and innovation programme under the \emph{Marie Sk\l odowska-Curie grant No 794592} and from the INdAM project \textit{Problemi isoperimetrici in spazi Euclidei e non}.
V.~F.\ and D.~P.\ also acknowledge the support of ANR-15-CE40-0018 project \textit{SRGI - Sub-Riemannian Geometry and Interactions}. B.~F.\ acknowledges the support of the Fondation Asile des Aveugles. 
\end{acknowledgements}
}

\appendix
\section{Orientation-dependent model of V1}\label{a:cortical}
%\todo{to read and correct}
Let us denote by $R>0$ the size of the visual plane and let $D_R\subset \bR^2$ be the disk $D_R:=\{x_1^2+x_2^2 \le R^2\}$. Fix $R>0$ such that $Q\subset D_R$.
In order to exploit the properties of the roto-translation group $SE(2)$ on images, we  now consider them to be elements of the set:
\begin{equation}
    \cI = \left\{f \in L^2(\bR^2,[0,1]) \text{ such that\ } \supp f\subset D_R\right\}.
\end{equation}
We remark that fixing $R>0$ is necessary, since contrast perception is strongly dependent on the scale of the features under consideration w.r.t.\ the visual plane. 

Orientation dependence of the visual stimulus is encoded via cortical inspired techniques, following e.g., \cite{Petitot,Citti2006,Duits2010,Prandi2017,Bohi2017}. 
The main idea at the base of these works goes back to the 1959 paper \cite{hubel1968receptive} by Hubel and Wiesel (Nobel prize in 1981) who discovered the so-called \emph{hypercolumn functional architecture} of the visual cortex V1.

Each neuron $\xi$ in V1 is assumed to be associated with a receptive field (RF) $\psi_\xi\in L^2(\bR^2)$ such that its response under a visual stimulus $f\in \mathcal I$ is given by 
\begin{equation}\label{eq:rf}
    F(\xi) = \langle \psi_\xi, f\rangle_{L^2(\bR^2)} = \int_{\bR^2} \overline{\psi_\xi({\color{black}z})} f({\color{black}z})\, d{\color{black}z}.
\end{equation}
Since each neuron is sensible to a preferred position and orientation in the visual plane, we let $\xi=(x,\theta)\in \mathcal{M} = \mathbb R^2\times \mathbb P^1$. Here, $\mathbb P^1$ is the projective line that we represent as $[0,\pi]/\sim$, with $0\sim \pi$. Moreover, in order to respect the \emph{shift-twist} symmetry \cite[Section ~4]{Bressloff2002}, we will assume that the RF of different neurons are ``deducible'' one from the other via a linear transformation. Let us explain this in detail.

The double covering of $\mathcal{M}$ is given by the Euclidean motion group $SE(2)=\bR^2\rtimes \bS^1$, that we consider endowed with its natural semi-direct product structure. That is, for $(x,\theta),(y,\varphi)\in SE(2)$, we let
\begin{gather}
  (x,\theta)\star(y,\varphi) 
  = 
 (x+R_\theta y, \theta+\varphi), 
% \quad
% \forall (x,\theta),(y,\varphi)\in SE(2), 
 \\
 \text{where }
 R_\theta = \left(\begin{array}{cc}
     \cos\theta & -\sin\theta \\
      \sin\theta & \cos\theta
 \end{array}\right).
\end{gather}
In particular, the above operation induces an action of $SE(2)$ on $\mathcal{M}$, which is thus an homogeneous space. %Indeed, identifying $\mathbb S^1 = [0,2\pi]/\sim$ and $\mathbb P^1 = [0,\pi]/\sim$, $SE(2)$ is the double covering of $M$ under the equivalence relation $\alpha\sim\beta \iff \alpha = -\beta \mod\pi$.
Observe that $SE(2)$ is unimodular and that its Haar measure (the %only 
left and right-invariant measure up to scalar multiples) is %simply 
$dxd\theta$.

We now denote by $\cU(L^2(\bR^2)) \subset \mathcal{L}(L^2(\bR^2))$ the space of linear unitary operators on $L^2(\bR^2)$ and let ${\color{black}\Pi}:SE(2)\to \cU(L^2(\bR^2))$   be the \emph{quasi-regular representation} of $SE(2)$. That is, ${\color{black}\Pi}(x,\theta)\in \cU(L^2(\bR^2))$ is the unitary operator encoding the action of the roto-translation $(x,\theta)\in SE(2)$ on square-integrable functions on $\mathbb R^2$. The action of ${\color{black}\Pi}(x,\theta)$ on $\psi\in L^2(\bR^2)$ is
\begin{equation*}
  [{\color{black}\Pi}(x,\theta)\psi](y) = \psi((x,\theta)^{-1}y)  = \psi(R_{-\theta}(y-x)), \quad \forall y\in \R^2.
\end{equation*}
Moreover, we let $\Lambda:SE(2)\to \cU(L^2(SE(2)))$ be the \emph{left-regular representation}, which acts on functions $F\in L^2(SE(2))$ as
\begin{equation}
  [\Lambda(x,\theta)F](y,\varphi) = F((x,\theta)^{-1}\star (y,\varphi)) 
%  = F(R_{-\theta}(y-x), \varphi-\theta)
  , \, \forall (y,{\color{black}\varphi})\in SE(2).
\end{equation}
% We will henceforth, with a little abuse of notation, denote by $\Lambda(x,\theta)$ also the element of  $\cU(L^2(\mathcal{M}))$ obtained by quotienting.

Letting %now 
$L:L^2(\bR^2)\to L^2(\mathcal{M})$ be the operator that transforms visual stimuli into cortical activations, one can formalise the \emph{shift-twist} symmetry by requiring %that
\begin{equation}
    L\circ {\color{black}\Pi}(x,\theta) = \Lambda(x,\theta)\circ L, \qquad \forall (x,\theta)\in SE(2).
\end{equation}
Under mild continuity assumption on $L$,  it has been shown in \cite{Prandi2017} that $L$ is then a continuous wavelet transform. That is, there exists a \emph{mother wavelet} $\Psi\in L^2(\bR^2)$ satisfying ${\color{black}\Pi}(x,\theta)\Psi = {\color{black}\Pi}(x,\theta+\pi)\Psi$ for all $(x,\theta)\in SE(2)$, and  such that
\begin{equation}\label{eq:wavelet}
Lf(x,\theta) = \langle {\color{black}\Pi}(x,\theta)\Psi, f \rangle, \quad \forall f\in L^2(\bR^2), (x,\theta)\in \mathcal{M}.
\end{equation}
Observe that the operation ${\color{black}\Pi}(x,\theta)\Psi$ above is well defined for $(x,\theta)\in \mathcal{M}$ thanks to the assumption on $\Psi$.
By \eqref{eq:rf}, the above representation of $L$ is equivalent to the fact that the RF associated with the neuron $(x,\theta)\in \mathcal{M}$ is the roto-translation of the mother wavelet, i.e., $\psi_{(x,\theta)}={\color{black}\Pi}(x,\theta)\Psi$.

\begin{remark}
  Letting $\Psi^*(x):=\overline{\Psi(-x)}$, the above formula can be rewritten as
\begin{equation}
  \begin{split}
    Lf(x,\theta) 
      &= \int_{\bR^2} \overline{\Psi(R_{-\theta}(y-x))}f(y)\,dy  \\
      &= \big[f * (\Psi^*\circ R_{-\theta})\big] (x),
      \quad \forall (x,\theta)\in SE(2).
  \end{split}
\end{equation}
where $f*g$ denotes the standard convolution on $L^2(\bR^2)$.
\end{remark}
%
%Notice that, although images are functions of $L^2(\bR^2)$ with values in $[0,1]$, it is in general not true that $Lf(x,\theta)\in [0,1]$. %However, one can check that by \eqref{eq:wavelet} it holds $|Lf|\le \|\Psi\|_{L^1(\bR^2)}$ whenever $f$ has values in $[0,1]$.
%However, from \eqref{eq:wavelet} we deduce the following result, which guarantees the boundedness of $L$. %is a bounded operator. 
%(See \cite[Section~2.2.3]{Prandi2017}.)
%
%\begin{prop}\label{prop:max-lift}
%  The operator $L:L^2(\bR^2)\to L^2(\mathcal{M})$ is continuous and, therefore, bounded.
%  In particular, for any $f\in L^2(\bR^2)$ the function $Lf$ is a continuous bounded function on $M$, with $|Lf|\le \|\Psi\|_{L^2(\bR^2)}\|f\|_{L^2(\bR^2)}$.
%  Moreover, if $f$ takes values in $[0,1]$ and $\Psi\in L^1(\bR^2)$, we have $|Lf|\le \|\Psi\|_{L^1(\bR^2)}$.
%\end{prop}

Neuro-physiological evidence shows that a good fit for the RFs is given by Gabor filters, whose Fourier transform is simply the product of a Gaussian with an oriented plane wave \cite{Daugman1985a}.  However, these filters are quite challenging to invert, and are parametrised on a bigger space than $\mathcal M$, which takes into account also the frequency of the plane wave and not only its orientation. For this reason, in this work we chose to consider as wavelets the \emph{cake wavelets} introduced in \cite{DuitsFelsbergetal2007}, see also \cite{Bekkers2014}. These are obtained via a mother wavelet $\Psi^{\text{cake}}$ whose support in the Fourier domain is concentrated on a fixed slice, which depends on the number of orientations one aims to consider in the numerical implementation. To recover integrability properties, the Fourier transform of this mother wavelet is then smoothly cut off via a low-pass filtering, see \cite[Section ~2.3]{Bekkers2014} for details. Observe, however, that in order to lift to $\mathcal M$ and not to $SE(2)$, we consider a non-oriented version of the mother wavelet, given by $\tilde\psi^{cake}(\mathbf{\omega}) + \tilde\psi^{cake}(e^{i\pi}\mathbf{\omega})$, in the notations of \cite{Bekkers2014}.

An important feature of cake wavelets is that, in order to recover the original image, it suffices to consider the \emph{projection operator} defined by
\begin{gather}
    P: L^2(\mathcal M)\to L^2(\bR^2), \\
    PF(x) := \int_{\mathbb P^1} F(x,\theta)\,d\theta,\qquad F\in L^2(\mathcal{ M})
\end{gather}
Indeed, by construction of cake wavelets, Fubini's Theorem shows that $(P\circ L)f = f$ for all $f\in \mathcal I$.

% Authors must disclose all relationships or interests that 
% could have direct or potential influence or impart bias on 
% the work: 
%
% \section*{Conflict of interest}
%
% The authors declare that they have no conflict of interest.

% BibTeX users please use one of
%\bibliographystyle{spbasic}      % basic style, author-year citations
\bibliographystyle{spmpsci}      % mathematics and physical sciences
\bibliography{biblio.bib}   % name your BibTeX data base

% Non-BibTeX users please use
% \begin{thebibliography}{}
% %
% % and use \bibitem to create references. Consult the Instructions
% % for authors for reference list style.
% %
% \bibitem{RefJ}
% % Format for Journal Reference
% Author, Article title, Journal, Volume, page numbers (year)
% % Format for books
% \bibitem{RefB}
% Author, Book title, page numbers. Publisher, place (year)
% % etc
% \end{thebibliography}

\end{document}